\documentclass[10pt,twocolumn,letterpaper]{article}

\usepackage{cvpr}
\usepackage{pdfpages}
\usepackage{multido}

\usepackage[accsupp]{axessibility} 

\usepackage{graphicx}
\usepackage{amsmath}
\usepackage{amssymb}
\usepackage{booktabs}
\usepackage{xspace}
\usepackage{times}
\usepackage{color}
\usepackage{float}
\usepackage{array}
\usepackage{graphicx}
\usepackage[percent]{overpic}

\usepackage[pagebackref,breaklinks,colorlinks]{hyperref}

\newcommand\blfootnote[1]{%
  \begingroup
  \renewcommand\thefootnote{}\footnote{#1}%
  \addtocounter{footnote}{-1}%
  \endgroup
}

\usepackage[capitalize]{cleveref}
\crefname{section}{Sec.}{Secs.}
\Crefname{section}{Section}{Sections}
\Crefname{table}{Table}{Tables}
\crefname{table}{Tab.}{Tabs.}

\def\cvprPaperID{8956} 
\def\confName{CVPR}

\definecolor{somegray}{rgb}{0.5, 0.5, 0.5}
\newcommand{\darkgrayed}[1]{\textcolor{somegray}{#1}}
\makeatletter
\newcommand*\titleheader[1]{\gdef\@titleheader{#1}}
\AtBeginDocument{%
  \let\st@red@title\@title
  \def\@title{%
    \vskip-3em
    \bgroup\normalfont\large\centering\@titleheader\par\egroup
    \vskip1.5em\st@red@title}
}
\makeatother
\titleheader{\darkgrayed{This paper has been accepted for publication at the \\
IEEE Conference on Computer Vision and Pattern Recognition (CVPR), New Orleans, 2022.
\copyright IEEE}}

\title{RGB-Multispectral Matching: Dataset, Learning Methodology, Evaluation}

\begin{document}

\author{ Fabio Tosi$^*$ \hspace*{1cm} Pierluigi Zama Ramirez$^*$  \hspace*{1cm} 
Matteo Poggi$^*$\\ Samuele Salti \hspace*{1cm} Stefano Mattoccia  \hspace*{1cm} Luigi Di Stefano \\
CVLAB, Department of Computer Science and Engineering (DISI)\\
University of Bologna, Italy\\
{\tt\small \{fabio.tosi5, pierluigi.zama, m.poggi\}@unibo.it}
}

\twocolumn[{
\renewcommand\twocolumn[1][]{#1}
\maketitle
\vspace{-1cm}
\begin{center}
    \centering
    \setlength{\tabcolsep}{3pt}
    \scalebox{0.7}{
    \begin{tabular}{cccc}

         \includegraphics[height=0.12\linewidth]{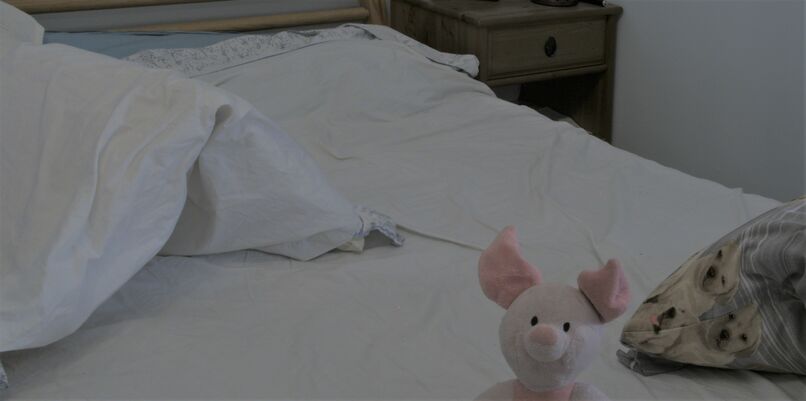} &
         \includegraphics[height=0.12\linewidth]{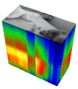} &
          \includegraphics[height=0.12\linewidth]{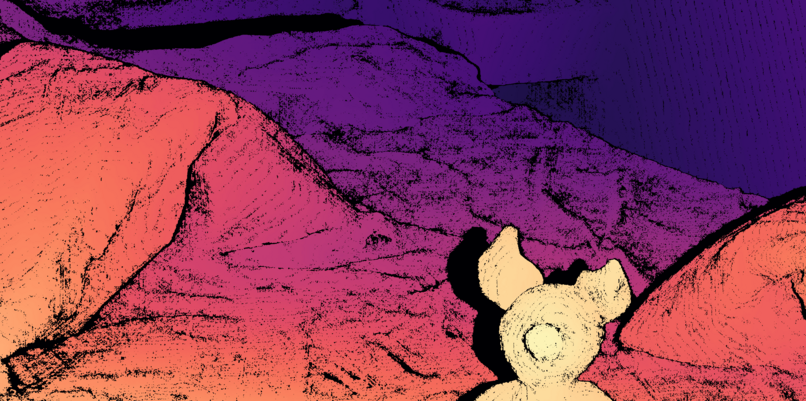} &
         \includegraphics[height=0.12\linewidth]{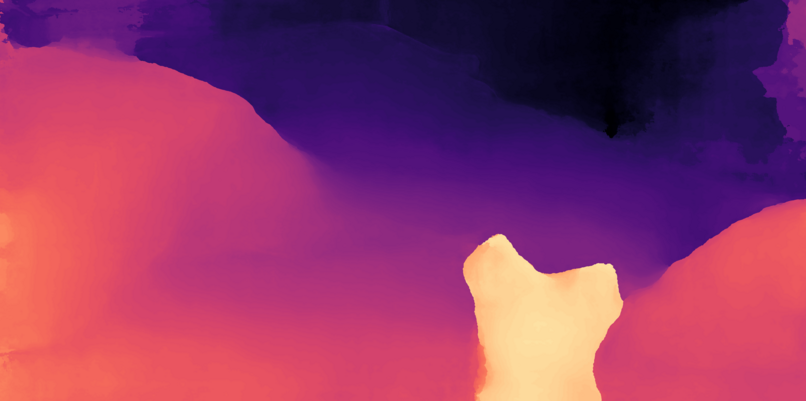} \\
         \includegraphics[height=0.12\linewidth]{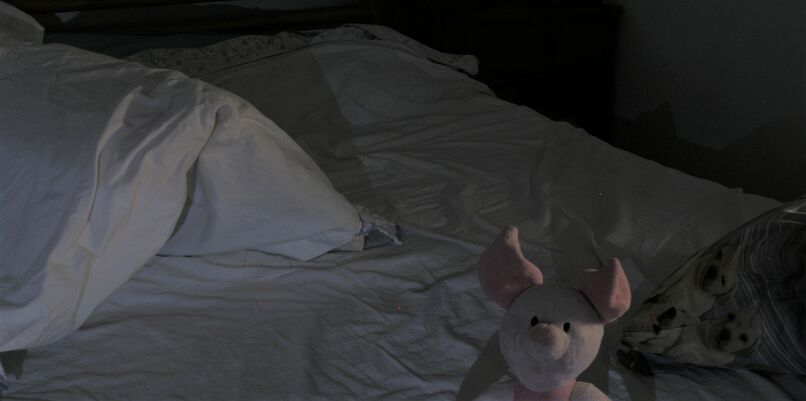} &
         \includegraphics[height=0.12\linewidth]{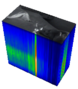} &
         \includegraphics[height=0.12\linewidth]{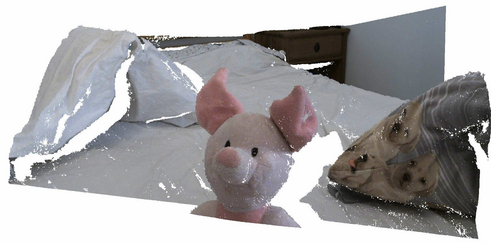} &
         \includegraphics[height=0.12\linewidth]{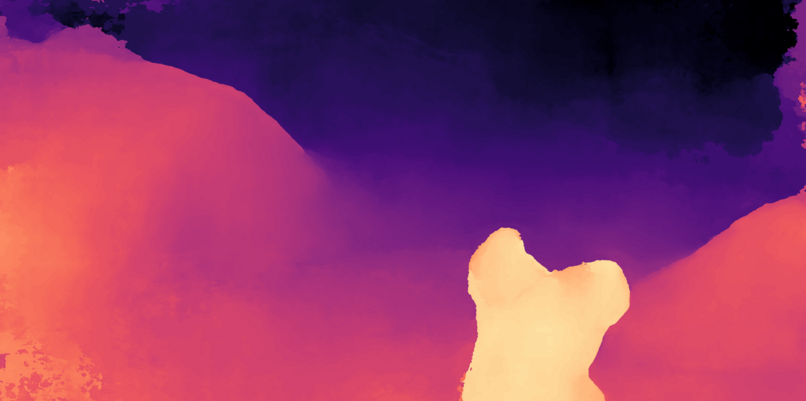} \\
         \textit{(a) RGB (3222$\times$1605$\times$3)} & \textit{(b) MS (510$\times$254$\times$10)} & \textit{(c) Ground-truth (3222$\times$1605)} & \textit{(d) Prediction (3222$\times$1605)} \\
    \end{tabular}}
    \label{fig:teaser}
\end{center}
\vspace{-0.2cm}
\small \hypertarget{fig:teaser}{Figure 1.} (a) RGB images and (b) low-resolution multi-spectral images acquired under varying lighting conditions. Visualization as a cube with on top the first band and on sides two stretched spectral slices of the bottom row and far-right column after the application of a color-map.
(c) ground-truth disparity aligned with the RGB image and its point-cloud visualization while (d)  prediction by our architecture.
\vspace{0.5cm}
}]

\maketitle

\begin{abstract}

We address the problem of registering synchronized color (RGB) and multi-spectral (MS) images featuring very different resolution by solving stereo matching correspondences. Purposely, we introduce a novel RGB-MS dataset framing 13 different scenes in indoor environments and providing a total of 34 image pairs annotated with semi-dense, high-resolution ground-truth labels in the form of disparity maps. To tackle the task, we propose a deep learning architecture trained in a self-supervised manner by exploiting a further RGB camera, required only during training data acquisition. In this setup, we can conveniently learn cross-modal matching in the absence of ground-truth labels by distilling knowledge from an easier RGB-RGB matching task based on a collection of about 11K unlabeled image triplets. Experiments show that the proposed pipeline sets a good performance bar (1.16 pixels average registration error) for future research on this novel, challenging task.

\end{abstract}

\section{Introduction}
\label{sec:intro}

\blfootnote{$^*$ Joint first authorship.}Traditional RGB sensors acquire images with three channels, approximately mimicking color perception in trichromatic mammals like humans, where three kinds of cone cells are sensitive to three different ranges of wavelengths in the visible spectrum. This is usually achieved by using bandpass optical filters corresponding to such colors, arranged in a Bayer pattern \cite{Bayer75patent}.
Multi-spectral (MS) imaging devices generalize such image acquisition mechanics and enable acquisition of images with a larger number of channels, i.e. ten or more, usually corresponding to narrower wavelength ranges. MS sensors may also be sensitive to wavelengths outside the visible spectrum, e.g.in the infra-red or ultra-violet bands. By extending the range of wavelengths as well as the granularity of their quantization into image channels, MS devices enable  extraction of additional information about the sensed scene that human eyes fail to capture, which in turns forms the basis of peculiar applications. 
For instance, MS imaging devices are used to perform analysis of artworks \cite{Colantoni06paintings}, remote sensing for agriculture \cite{Zhou18Crops} and land use \cite{Zhong18land}, target tracking \cite{Ge19target}, pedestrian detection \cite{Hwang15MSPedestrian}, counterfeit detection, e.g. of banknotes \cite{BAEK18banknotes}, diagnostic medicine and skin inspection \cite{Kuzmina11skin}, food inspection \cite{QIN13food} and contamination detection \cite{Garavelli17FoodContaminant}. In spite of such broad range of applications, deployment of MS sensors is usually limited to industrial settings or expensive equipment like satellites. However, the ability to run several of these applications on mobile phones and other consumer devices directly operated by end users could open up interesting scenarios, like, e.g., enabling early diagnosis of skin diseases \cite{Kim16smartphone}, democratizing food quality control \cite{Bandara18Multispectral}, making plant phenotyping easier \cite{WANG20LeafSpec}, and others yet to imagine. 

While the number of traditional cameras on high-end phones have been growing steadily in recent years, MS sensors have not been ported yet on consumer devices like phones or action cameras due to several limitations \cite{Genser20MSArray}. In particular,  MS sensors resolution is significantly smaller than standard RGB cameras which feature several Megapixels of resolution, because the most suitable technology to realize them extends the Bayer pattern used for color imaging into multi-spectral filter arrays \cite{Lapray14MSFA}, with each native pixel of the imaging sensor detecting one band by placing in front of it the corresponding optical filter, i.e. one MS ``pixel" for a camera sensing 16 bands uses a 4$\times$4 grid of native pixels. Thus, MS sensors tend also to be larger and bulkier than RGB cameras; and they are orders of magnitude more expensive, \ie MS cameras cost at least tens of thousands of dollars/euros. There exist linear MS cameras \cite{Tack12LinearMS} or MS cameras realized with filter wheel technology \cite{Brauers08FWMScamera} which may feature high resolution but can sense only static scenes and are not appropriate for deployment on mobile devices.

As a result of these technological limitations, MS cameras compatible with cost and size requirements of mobile devices feature very small resolution, insufficient for the applications listed above. Moreover, beside up-sampling the MS image to usable resolutions, it is usually important to align MS information with RGB streams coming from traditional cameras on the device, where objects or areas of interest can be easily identified with effective algorithms. Due to the challenges of matching images across spectra, the mostly explored setup to acquire  MS images and RGB images simultaneously has been to physically align their optical centers by using beam splitters \cite{Cao11HighResMS,Hwang15MSPedestrian}, which are however unfeasible to deploy in mobile devices, where instead the dense registration between the images in the pair must be computed by computer vision algorithms.

Although establishing dense image correspondences is one of the fundamental and most studied problems in computer vision, solutions that address the cross-spectral images and/or unbalanced resolutions  are rare. They only investigated case is the special incarnation of the problem where an RGB image is matched to a Near Infra-Red (NIR) or Infra-Red (IR) one at the same resolution \cite{chiu2011improving,shen2014multi,kim2016dasc,kim2016deep,zhi2018deep}. To the best of our knowledge, the general MS-RGB case, both balanced and unbalanced in terms of resolution, is yet unexplored in literature. Research on this topic has also been hindered by the lack of publicly available datasets: existing methods tackling the NIR/IR-RGB case have been tested on datasets acquired with custom hardware setups targeting specific use cases, like autonomous driving or object detection, and never made publicly available \cite{chiu2011improving,zhi2018deep} or are tested on small datasets with sparse ground-truth: e.g. the dataset used by \cite{kim2016dasc,kim2016deep} and proposed in \cite{shen2014multi} has 7 image pairs where 50-100 object corners on average per pair have been manually annotated, i.e. less 700 ground-truth correspondences. 

In this work, we propose the first large-scale and publicly-available dataset to study RGB-MS registration. 
In particular, we cast the registration problem as a \textit{stereo matching} one due to the two cameras being synchronized. 
Our dataset features more than 11K stereo pairs composed of a low-res 510$\times$254 MS image and a high-res 3222$\times$1605 RGB image which can be used to compute a high-res MS image registered at pixel-level with the RGB image. Examples of pairs are shown in Fig.\ \hyperref[fig:teaser]{1}a-b. Another key feature of our dataset is that 34 stereo pairs coming from 13 different scenes are densely annotated (Fig.\ \hyperref[fig:teaser]{1}c), thanks to an original acquisition methodology whereby a second RGB image and several projectors are used to create a very accurate active space-time stereo setup \cite{davis2003spacetime}, which result in more than 125 millions of ground-truth correspondences. We also propose a deep learning architecture to tackle the challenging cross-spectral and resolution-unbalanced problem, that can be used to compute registered MS images at arbitrary resolutions and serves as a baseline for the dataset, whose results are shown in Fig.\ \hyperref[fig:teaser]{1}d. When training the network, we leverage the large body of unlabelled images in our dataset by sourcing \textit{proxy}-labels, which are obtained by exploiting again the second RGB camera to run passive stereo matching. Our dataset enables the community to study the challenging problem of cross-spectral and resolution-unbalanced matching of images, which is key to enable porting of existing MS applications in the consumer space as well as to unlock new applications of MS imaging specific to the mobile device setup. Our contributions are:

$\bullet$ we propose the first investigation into the challenging problem of cross-spectral and resolution-unbalanced dense matching, which is a key enabling technology to unlock MS applications on mobile devices;

$\bullet$ we present the first large-scale publicly available dataset for the problem;

$\bullet$ thanks to a peculiar acquisition methodology exploiting two registered high-res RGB cameras, we also make available the first densely labelled set of images for this problem and we propose a training methodology which can leverage unlabelled images via proxy supervision;

$\bullet$ we propose a deep architecture to compute correspondences between images at different resolutions and with different spectral content, which can be used to generate MS images registered at pixel-level with the RGB stream.

The project page is available at \url{https://cvlab-unibo.github.io/rgb-ms-web/}.

\section{Related Work}
\textbf{Deep Stereo.} First attempts to learn in stereo matching focused on the design of robust matching functions between image patches, implemented with shallow CNNs \cite{zbontar2016stereo,Chen_2015_ICCV,luo2016efficient}. Then, along the established pipeline \cite{scharstein2002taxonomy} other sub-tasks were cast in the form of a neural network, such as optimization \cite{seki2016patch,seki2017sgm-net} and disparity refinement \cite{gidaris2017detect,Jie_2018_CVPR,batsos2018recresnet,ferrera2019fast,aleotti2021neural}. However, the development of end-to-end architectures \cite{mayer2016large}
represented a real turning point in the field, with more and more works focusing nowadays on the design of new architectures rather than on classical algorithms.
According to \cite{poggi2021synergies}, two main families of deep stereo networks exist, respectively 2D \cite{mayer2016large,Pang_2017_ICCV_Workshops,Liang_2018_CVPR,song2018edgestereo,Ilg_2018_ECCV,yang2018segstereo,lipson2021raft} and 3D architectures \cite{Kendall_2017_ICCV,chang2018psmnet,yang2019hierarchical,dovesi2020real,zhang2019ga,zhang2019domaininvariant,shen2021cfnet,cai2020matchingspace}.

\textbf{Self/Proxy-Supervised Stereo.} In parallel to the spread of end-to-end stereo, some first strategies were developed to train such models without ground-truth labels \cite{Zhou_2017_ICCV,Tonioni_2017_ICCV}. Several works followed, exploiting image reprojection across the two views as a form of supervision \cite{Tonioni_2019_CVPR,wang2019unos,lai19cvpr} or, in alternative, the guidance given by traditional stereo algorithms \cite{Tonioni_2017_ICCV,tonioni2020unsupervised,Poggi2021continual,aleotti2020reversing} was used to distill \textit{proxy} labels to supervise the stereo network.

\textbf{Cross-spectral Matching}. 
Finding correspondences between images sensing different parts of the wavelength spectrum represents an additional challenge. This task has been investigated in different settings, in most cases by matching RGB-IR \cite{chiu2011improving,mehltretter2018multimodal}, RGB-thermal \cite{pinggera122012cross} and RGB-NIR \cite{zhi2018deep,shen2014multi,kim2016dasc,kim2016deep} modalities, as well as between stereo images with very different radiometric variations \cite{heo2010robust,heo2012joint}. Traditionally, most approaches aimed at designing hand-crafted functions \cite{heide2018real,chiu2011improving,shen2014multi,heo2010robust} or descriptors \cite{pinggera122012cross,kim2016dasc,heo2012joint}. 
More modern approaches deployed deep-learning to learn a deep correlation function \cite{kim2016deep} starting from a self-similarity measure while
Zhi \etal \cite{zhi2018deep} designed an end-to-end network trained in semi-supervised manner on RGB-NIR, by explicitly leveraging knowledge about materials after manual annotation of the training set. 

Only a few recent works \cite{9043847, 8575758, 9554517} on remote sensing focus on affine trasformations to tackle the RGB-MS registration task. However, in contrast to our method, they do not explicitly reason on searching dense correspondences between the two input signals. Among published papers, \cite{zhi2018deep} is the most relevant to ours, as it proposes an end-to-end model trained without ground-truth, although in the RGB-NIR setup.
Conversely to \cite{zhi2018deep}, our novel \textit{proxy}-supervised training paradigm does not require any manual annotation, conveniently exploiting an additional RGB camera during training data acquisition.
Moreover, we propose a novel dataset with semi-dense ground-truth disparity maps, counting a total of more than 125M annotated pixels, whereas in \cite{zhi2018deep} were limited to a few, sparse pixels, \ie 5K.

\section{RGB-MS Dataset}

In this section, we present our novel RGB-MS dataset. We start by describing the acquisition setup and then we dig into the procedure to annotate the RGB-MS pairs with semi-dense ground-truth labels.

\subsection{Camera Setup}

Our acquisition setup consists of two synchronized Ximea cameras: an RGB camera equipped with a Sony IMX253LQR-C 12.4 Mpx sensor and an MS camera based on a IM-SM4X4-VIS2 2.2 Mpx sensor with 10 bands. Both cameras are mounted on a common support with a distance of about 4cm between their optical centers. 

The two cameras are calibrated, so as to form a stereo system where the RGB device (left) is used as the reference camera. This is achieved  by the  OpenCV tools for camera calibration \cite{opencv_library}, after having acquired a set of images with both cameras framing a planar chessboard. Then, corner detection is performed on both images after conversion to grayscale. As for the  MS camera, we found it sufficient to average the 10 bands pixel-wise in order to obtain a pseudo-grayscale image in which corners are easily detectable. Thus, we estimate intrinsic and lens distortion parameters for both cameras and then, after undistortion, calibrate the RGB-MS stereo camera system and estimate the transformations to rectify the RGB and MS frames. 
Given the dramatic difference between the two images in terms of resolution, we cannot rely on the standard procedure to estimate the rectification parameters, since it would lead to rectified images resized to a resolution in between the two. 
Thus, we define the concept of unbalance rectification in which we consider two images as \textit{unbalance rectified} if, after resizing them to same resolution, they are rectified. Additional details in the supplementary material.

\subsection{Annotation Pipeline}

We now introduce our annotation pipeline, aimed at enriching our RGB-MS dataset with semi-dense ground-truth disparity maps. A variety of active depth sensors would fit this purpose, although, on the one hand, significantly limiting the resolution at which our ground-truth could be collected, on the other introducing the need to carry out a registration between the RGB camera and the active sensor itself, a non-trivial task itself.

To overcome these issues, we leverage a space-time stereo pipeline \cite{davis2003spacetime} by adding a second RGB sensor to our camera setup. Specifically, we mount this additional device on the right of the MS sensor, i.e. with a baseline of about 8cm with respect to the other RGB camera, thus implementing a 12.2 Mpx RGB stereo setup. This configuration allows to fully exploit the high resolution of the RGB cameras to provide much more accurate ground-truth labels compared to the deployment of any ancillary active sensor (e.g. Lidar or ToF) while removing the need of complex registration across sensors. Moreover, the larger baseline increases depth resolution and produces finer-grained labels.

\begin{figure}
    \centering
    \setlength{\tabcolsep}{2pt}
    \begin{tabular}{ccc}
        \small{\textit{Active Stereo}} & \small{\textit{Pre-Warping}} & \small{\textit{Warped}} \\
        \begin{overpic}[width=0.31\linewidth]{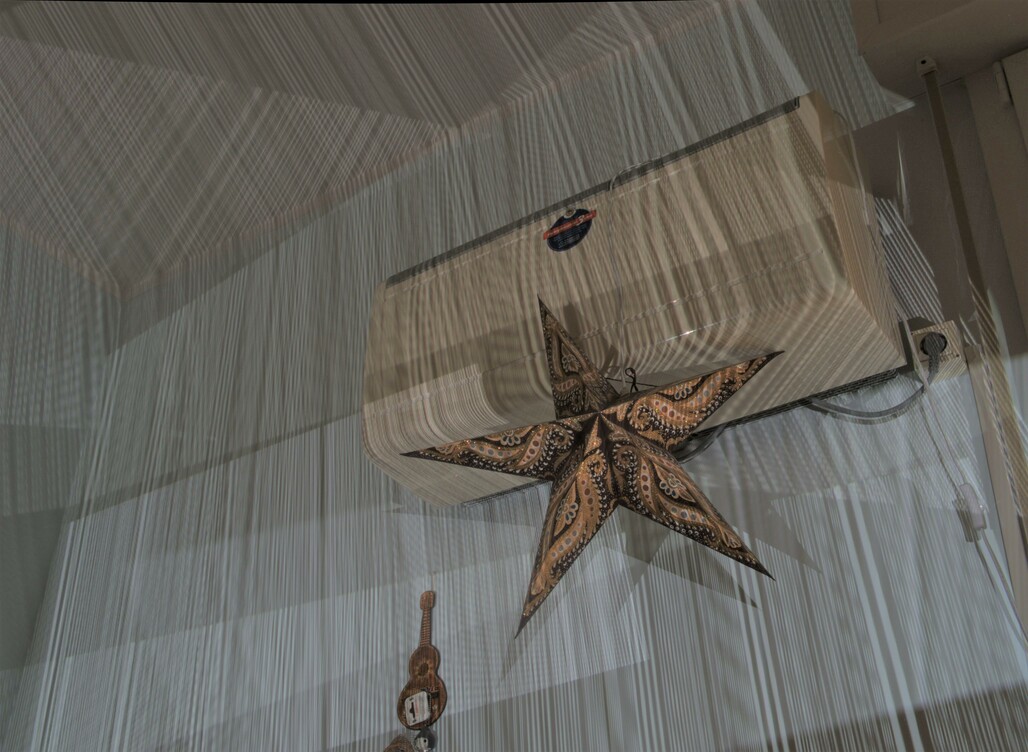} 
        {\colorbox{white}{$\displaystyle\textcolor{black}{\text{(a)}}$}}
        \end{overpic}
        &
        \begin{overpic}[width=0.31\linewidth]{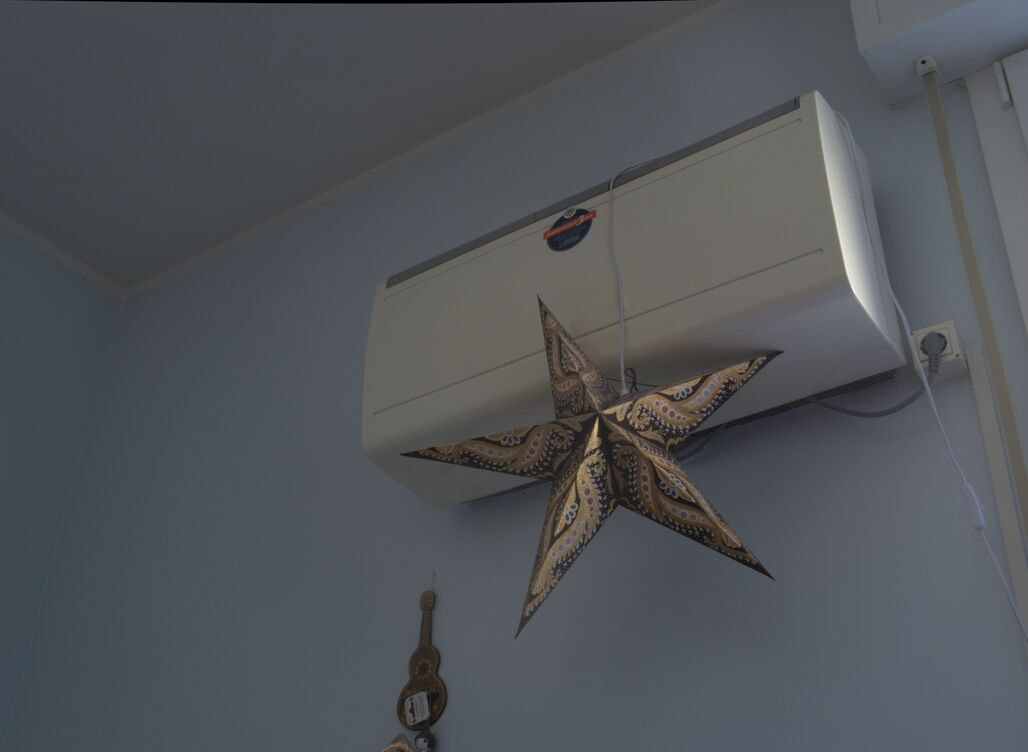} 
        {\colorbox{white}{$\displaystyle\textcolor{black}{\text{(c)}}$}}
        \end{overpic}
        &
        \begin{overpic}[width=0.31\linewidth]{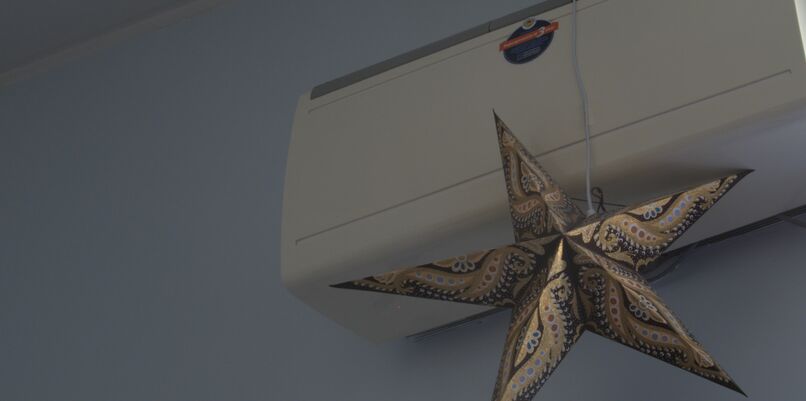} 
        {\colorbox{white}{$\displaystyle\textcolor{black}{\text{(e)}}$}}
        \end{overpic} \\
        \begin{overpic}[width=0.31\linewidth]{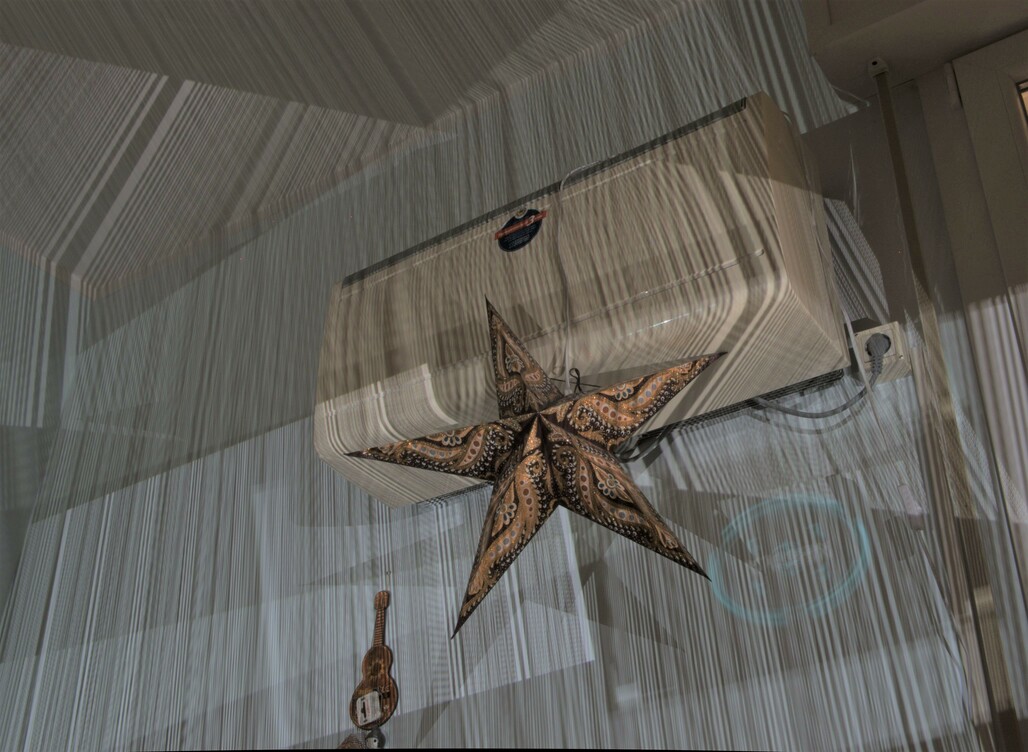} 
        {\colorbox{white}{$\displaystyle\textcolor{black}{\text{(b)}}$}}
        \end{overpic}
        &
        \begin{overpic}[width=0.31\linewidth]{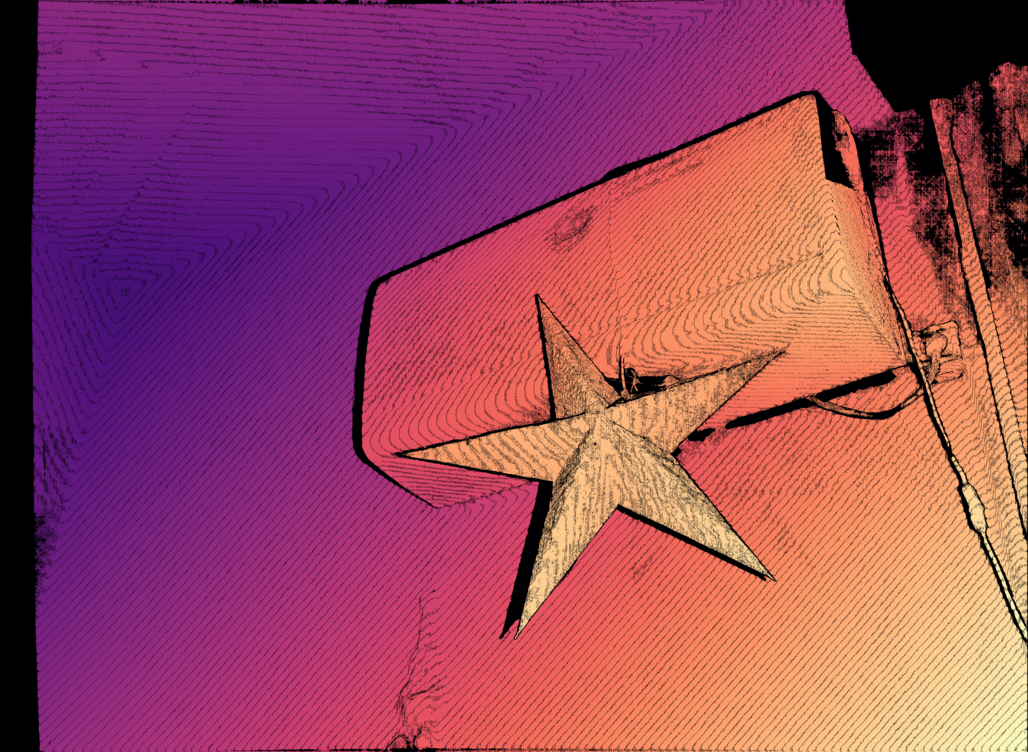} 
        {\colorbox{white}{$\displaystyle\textcolor{black}{\text{(d)}}$}}
        \end{overpic}
        &
        \begin{overpic}[width=0.31\linewidth]{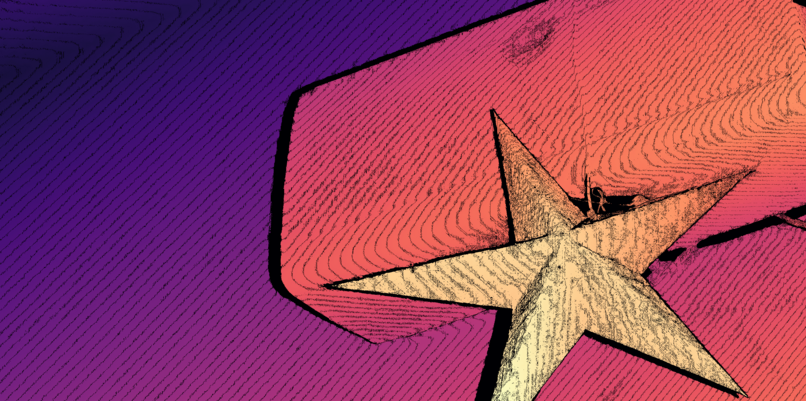}
        {\colorbox{white}{$\displaystyle\textcolor{black}{\text{(f)}}$}}
        \end{overpic} \\
        4112$\times$3008$\times$3 & 4112$\times$3008$\times$3 &
        3222$\times$1605$\times$3 \\
    \end{tabular}
    \vspace{-0.2cm}\caption{\textbf{Ground-truth Acquisition.} Given a set of active RGB-RGB stereo pairs (a,b), we compute the ground-truth disparity (d) aligned with the left image of the RGB-RGB stereo system (c) with our space-time stereo algorithm. Then, we warp it (f) to be aligned with the left image of the RGB-MS stereo system (e).}
    \label{fig:dataset_pipeline}
\end{figure}

\textbf{Image Acquisition.} To collect a single scene for which we aim at providing an RGB-MS pair together with ground-truth disparities, we use our trinocular cameras setup to acquire a set of \textit{passive} RGB-MS images, \ie the static scene is acquired in absence of any external perturbation. For each scene, we make several acquisition with different illuminations. These frames represent the actual images that will be distributed with our dataset. 
Then, we use a set of portable projectors to perturb the sensed environment with random black-and-white banded patterns generated through Non-recurring De Bruijn sequence \cite{lim2009optimized}, as shown in Fig.\ \ref{fig:dataset_pipeline}a-b. We acquire several dozens of \textit{active} images for each scene to be processed by the next step in the pipeline.

\textbf{Space-Time Semi-Global-Matching.} We implement a robust space-time stereo pipeline \cite{davis2003spacetime} to process the active stereo pairs acquired on the same static scene and obtain a highly accurate disparity map (Fig.\ \ref{fig:dataset_pipeline}d). Such a framework leverages on the variety of patterns projected during each acquisition, greatly increasing the distinctiveness of any single pixel and thus easing the matching process. In particular, our algorithm implements four steps:

i) \textit{Single-pair cost computation.} For each active RGB-RGB pair $t$, we compute an initial Disparity Space Image (DSI), storing $d_{max}$ matching costs for any pixel in the reference image at coordinates $(x,y)$. Such costs are obtained by applying a Census transform \cite{zabih1994non} on $9\times7$ windows to both frames and then computing the Hamming distance between  63-bit strings:
\begin{equation}
    \text{DSI}_t(x,y,d) = \sum_i \mathcal{C}_i^{L(t)}(x,y) \oplus \mathcal{C}_i^{R(t)}(x-d,y)
\end{equation}
with $\mathcal{C}^{L(t)}, \mathcal{C}^{R(t)}$ being the $t$-th left and right census transformed images and $i$ any single bit in the 63-bit strings.

ii) \textit{Space-time integration and SGM optimization.} Once single DSI$_t$ volumes have been initialized, we integrate them over time to obtain more robust matching costs thanks to the variety of different patterns projected onto the scene:

\begin{equation}
    \text{DSI}(x,y,d) = \sum_t \text{DSI}_t(x,y,d)
\end{equation}
Then, we further optimize the matching cost distributions by means of the Semi-Global Matching algorithm \cite{hirschmuller2007evaluation}, before selecting the final disparity $\hat{d}(x,y)$ by means of a Winner-Takes-All strategy. 

iii) \textit{Outliers suppression.} Although SGM dramatically regularizes the initial cost volume and leads to smooth disparity maps, several outliers are still present.  
To remove them, we apply a confidence-based approach \cite{tosi2017learning} to filter out the least confident pixels. Given a pool of conventional confidence measures, we discard all pixels marked as not reliable according to each measure. To this aim, we select two binary measures from \cite{poggi2021confidence} -- ACC, LRC -- together with a Weighted Median Disparity Deviation over a $41\times 41$ window, considering pixels as not confident if larger than 1.
    
iv) \textit{Sub-pixel interpolation.} Finally, we estimate subpixel disparities to improve depth resolution. Starting from the DSI output of SGM, an interpolation algorithm based on \cite{miclea2015new} is applied to pixels that have not been filtered out previously. This is traditionally achieved by fitting, for any pixel $(x,y)$ in the image, a function between its minimum cost $c_{\hat{d}}=DSI(x,y,\hat{d})$ and its disparity neighbours $c_{\hat{d}-1}$, $c_{\hat{d}+1}$. The interpolation function can be assumed as monotonically increasing in $[0,1]$, then defining sub-pixel precise disparity $\hat{d}_\text{sub}$ as
\begin{equation}
    \hat{d}_\text{sub} =
    \begin{cases}
        \hat{d} - 0.5 + \frac{c_{\hat{d}-1} - c_{\hat{d}}}{c_{\hat{d}+1} - c_{\hat{d}}} &  \textit{if} \; c_{\hat{d}-1} > c_{\hat{d}+1} \\
        \hat{d} - 0.5 + \frac{c_{\hat{d}+1} - c_{\hat{d}}}{c_{\hat{d}-1} - c_{\hat{d}}} & \text{otherwise} \\
    \end{cases}
\end{equation}
Following \cite{miclea2015new}, we fit a low-order polymonial function, \ie a parabola $ax^2+bx+c$, setting $a=1$, $b=1$ and $c=0$.

\textbf{Manual Cleaning on Point Clouds.} Despite the aforementioned filtering strategies, the disparity maps obtained so far may still present some outliers. Thus, we manually clean the ground-truth maps of any noisy label by projecting them into 3D point clouds and selecting all points resulting isolated from the main structures in the scene. For pixels corresponding to points that have been filtered out in the point cloud, we remove them from the ground-truth maps.

\textbf{Disparity Warping.}
The  ground-truth disparity maps produced so far are aligned to the reference image of the RGB-RGB setup, i.e. the left image rectified for the RGB-RGB setup (Fig.\ \ref{fig:dataset_pipeline}c). The reference image for the RGB-MS setup is the same, but it is rectified differently due to the different mechanical alignment and the smaller field of view of the MS camera (see supplementary for more details).
Hence, we use the estimated rectification homographies to perform backward warping of the disparity map and align it with the RGB image rectified for the RGB-MS setup (Fig.\ \ref{fig:dataset_pipeline}e), while considering the rotation of the camera reference frame and the different baselines in the RGB-MS and RGB-RGB setups, thus producing the final ground-truth disparity maps for the RGB-MS pair (Fig.\ \ref{fig:dataset_pipeline}f).

\section{Cross-Spectral Matching} 

\begin{figure}
    \centering
    \includegraphics[trim={7cm, 2.8cm, 9cm, 0.5cm}, clip, width=0.8\linewidth]{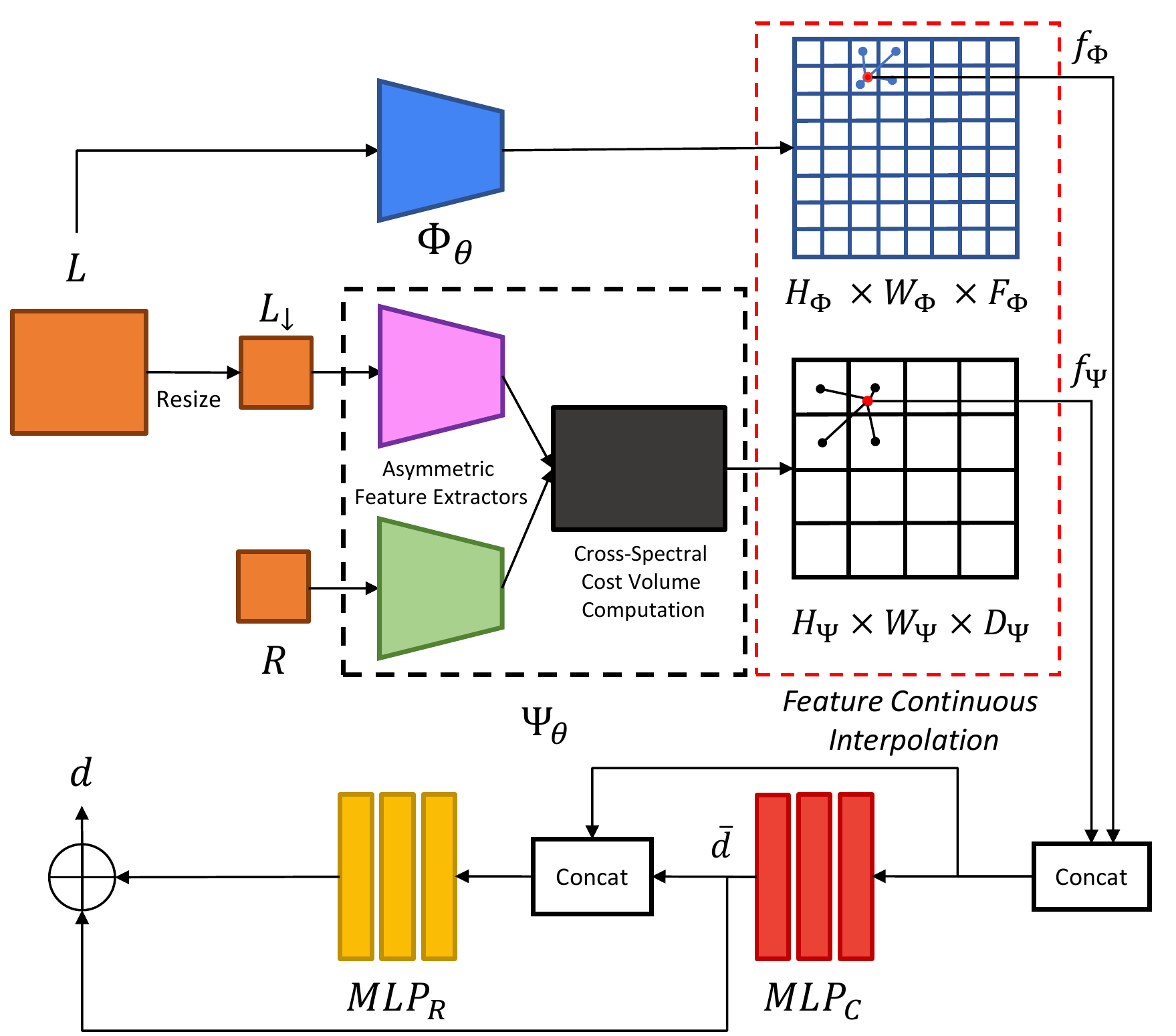}
    \vspace{-0.2cm}\caption{\textbf{Architecture Overview.} Given an unbalanced stereo pair composed of a reference high-resolution image $L$ and a target multi-spectral low-resolution image $R$, our network estimates a disparity map aligned with $L$ by combining cross-spectral cost probabilities computed by a stereo backbone $\Psi_\theta$ and deep features from $L$ obtained by the feature extractor $\Phi_\theta$ . } 
    \label{fig:network}
\end{figure}

Fig. \ref{fig:network} illustrates our deep architecture specifically designed to address the RGB-MS matching problem. 
Given a rectified RGB-MS pair, we extract deep features from the high-resolution RGB image and compute the matching cost volume between the MS image and the RGB image resized at the same spatial resolution. Then, the extracted features and the cost volume are used to estimate a continuous dense correspondence field, which can be used to synthesize a high-resolution multi-spectral image aligned with the RGB one. In this section, we describe in detail the network architecture, the loss function and the training protocol.

\subsection{Problem Statement}

Let $L \in \mathbb{R}^{W_L \times H_L \times 3}$  denote the RGB image acquired by the high-resolution camera and $R \in \mathbb{R}^{W_R \times H_R \times N}$ ($N=10$ in our experiments) the image captured by the low-resolution MS sensor after unbalanced rectification. As highlighted in Fig. \hyperref[fig:teaser]{1}, the two cameras feature very different properties in terms of resolution and number of channels. Our goal is to estimate a disparity map $\mathcal{D}$ aligned with the reference image $L$ at an arbitrary spatial resolution.

\subsection{Deep Cross-Spectral Network}

Our proposed architecture composed of 1) a coarse sub-module responsible for integrating geometric information between the RGB image and the MS signal at low-resolution and 2) a fine-grained sub-module aimed at recovering details by leveraging on the original high-resolution RGB image. 
More specifically, let 
\begin{equation}
\Psi_\theta: \mathbb{R}^{W_R \times H_R \times (N+3)} \rightarrow \mathbb{R}^{W_{\Psi} \times H_{\Psi} \times D_{\Psi}}
\end{equation}
denote a deep stereo backbone with parameters $\theta$ and maximum disparity $D_{\Psi}$. Examples for such backbones are stereo architectures that process a standard stereo pair by means of siamese towers and perform 3D convolutions on 4D feature volumes \cite{chang2018psmnet,guo2019group,yang2019hierarchical,zhang2019ga} to produce a 3D cost volume without performing the last softargmax step. For our specific setup, $\Psi_\theta$ takes as input $L_\downarrow$ and $R$, where $L_\downarrow$ represents the reference RGB image $L$ resized to match the resolution of $R$, and outputs a feature vector $D_{\Psi}$. By doing so, the two images can be considered rectified based on the  definition of unbalanced rectification \cite{aleotti2021neural}. However, differently from the standard RGB-RGB setting where the stereo pair is processed by means of siamese towers, we modify $\Psi_\theta$ to compute deep features from $L_\downarrow$ and $R$ using two feature extractors -- that are independent, because of the different information sensed in the two images and encoded in a different number of input channels ($L_\downarrow$:3, $R$: N). Once high-dimensional features have been extracted from the two images, we keep unchanged the remaining part of the stereo backbone $\Psi_\theta$ in charge of building the  cross-spectral cost volume and regularizing the aggregated costs. Our intuition is that, by doing so, the network effectively learns to match deep features computed from two completely different signals at low-resolution and, thus, to recover the final 3D structure of the scene.  Yet, $\Psi_\theta$ does not make full use of the available high resolution RGB image, where important high-frequency details may be extracted. Thus, we deploy also image features learned from the high-resolution image by means of a fully convolutional module that has the same architecture of the low resolution feature extractor adopted in $\Psi_\theta$ but does not share weights with it and defined as:
\begin{equation}
\Phi_\theta: \mathbb{R}^{W_L \times H_L \times 3} \rightarrow \mathbb{R}^{W_{\Phi} \times H_{\Phi} \times F_{\Phi}}
\end{equation}
where $F_{\Phi}$ represents the dimension of extracted features.

Following recent advances in continuous function representations \cite{Tosi2021CVPR, aleotti2021neural}, we adopt a Multilayer Perceptron (MLP) to compute the final disparity map $d \in \mathbb{R}$ at arbitrary spatial location  $\mathbf{x}_L \in \mathbb{R}^2$ where the input features for the MLP are obtained by concatenating the bilinear interpolated features $f_{\Psi}$ and $f_{\Phi}$ computed on the output of $\Psi_\theta$ and $\Phi_\theta$, respectively.
We train our model by adopting the recent loss function proposed in \cite{aleotti2021neural} that allows estimating accurate disparity maps as well as sharp depth boundaries. 
More specifically, we use two MLPs: MLP$_C$, in charge of computing a categorical distribution over disparity values $[0,d_\text{max}]$ from which an initial integer disparity value is selected, and MLP$_R$, which estimates a real-valued offset to recover  subpixel information. The final disparity value predicted by our network can be expressed as: 
\begin{equation}
         d = \bar{d} + \text{MLP}_R(f_{\Psi}(\mathbf{x}_{L_\downarrow}), f_{\Phi}(\mathbf{x}_L), \bar{d})
\end{equation}
where $\bar{d}$ is the integer disparity computed by MLP$_C$, i.e.
\begin{equation}
         \bar{d} = \textrm{argmax}(\text{MLP}_C(f_{\Psi}(\mathbf{x}_{L_\downarrow}), f_{\Phi}(\mathbf{x}_L)))
\end{equation}

where $\mathbf{x}_{L_\downarrow} \in \mathbb{R}^2$ is the 2D continuous location in $L_\downarrow$, while $\mathbf{x}_L \in \mathbb{R}^2$ is the corresponding location at high resolution, defined as $\mathbf{x}_L = \frac{W_L}{W_{L_\downarrow}}\mathbf{x}_{L_\downarrow}$. 

Thus, by indicating the disparity labels as $d^*$, our network can be trained by minimizing the following loss function:
\begin{equation} \label{eq:loss}
    \begin{split}
        \mathcal{L} = &-\mathcal{N}(d^*,\sigma) * \textrm{log}\Big( \text{MLP}_C(f_{\Psi}(\mathbf{x}_{L_\downarrow}), f_{\Phi}(\mathbf{x}_L)) \Big) \\
        &+ \Big| MLP_R(f_{\Psi}(\mathbf{x}_{L_\downarrow}), f_{\Phi}(\mathbf{x}_L)) -d^*_R \Big| \\
    \end{split}
\end{equation}
where $\mathcal{N}(d^*,\sigma)$ represents the Gaussian distribution centered at $d^*$ and sampled at integer values in $[0,d_\text{max}]$, while $d^*_R=d^* - \bar{d}$.

\begin{figure}
    \centering
    \setlength{\tabcolsep}{2pt}
    \begin{tabular}{ccc}
        \small{\textit{RGB}} & \small{\textit{MS}} & \small{\textit{Proxies}} \\
         \begin{overpic}[width=0.31\linewidth]{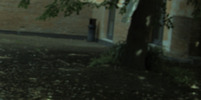}
         {\colorbox{white}{$\displaystyle\textcolor{black}{\text{(a)}}$}}
         \end{overpic} &
        \begin{overpic}[width=0.31\linewidth]{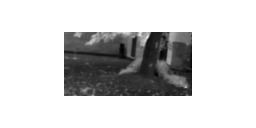} 
        {\colorbox{white}{$\displaystyle\textcolor{black}{\text{(b)}}$}}
        \end{overpic}&
        \begin{overpic}[width=0.31\linewidth]{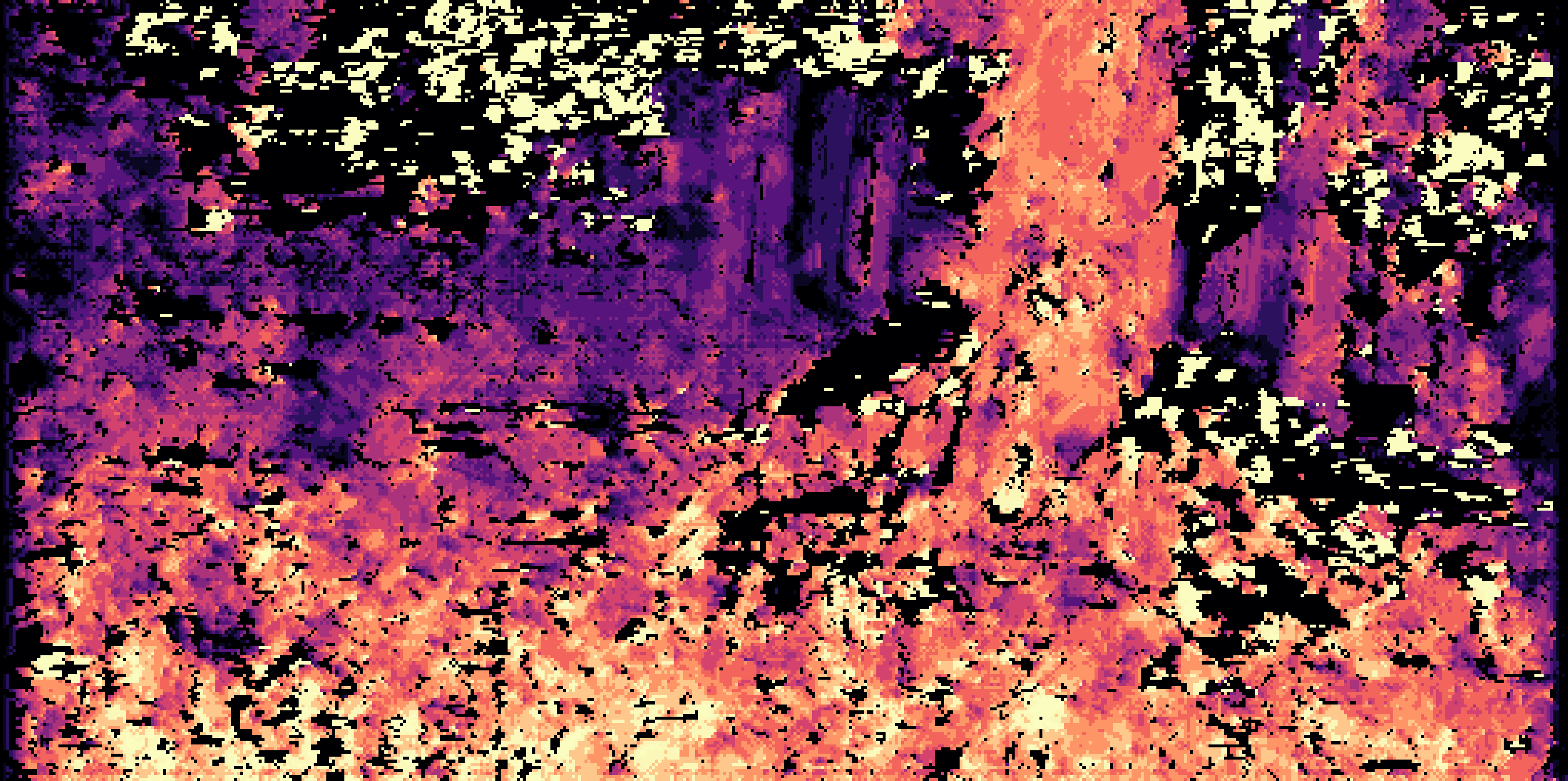} 
        {\colorbox{white}{$\displaystyle\textcolor{black}{\text{(c)}}$}}
        \end{overpic}\\
        \small{\textit{RGB}} & \small{\textit{2$^\circ$ RGB}} & \small{\textit{Warped Proxies}} \\
        \begin{overpic}[width=0.31\linewidth]{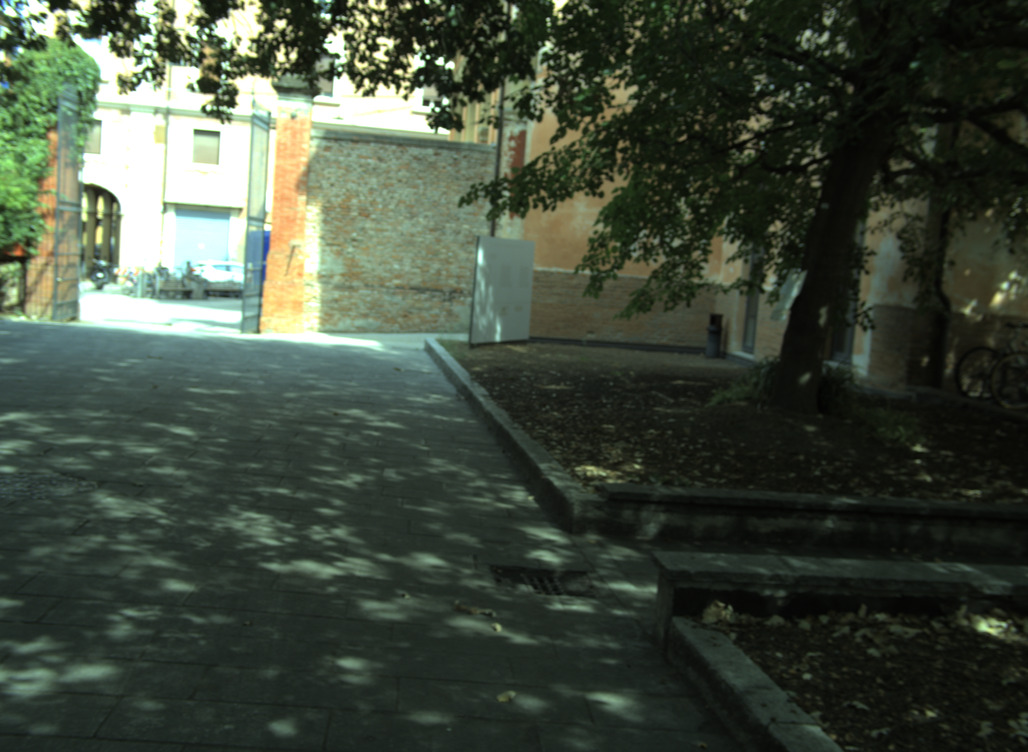} 
        {\colorbox{white}{$\displaystyle\textcolor{black}{\text{(d)}}$}}
        \end{overpic} &
        \begin{overpic}[width=0.31\linewidth]{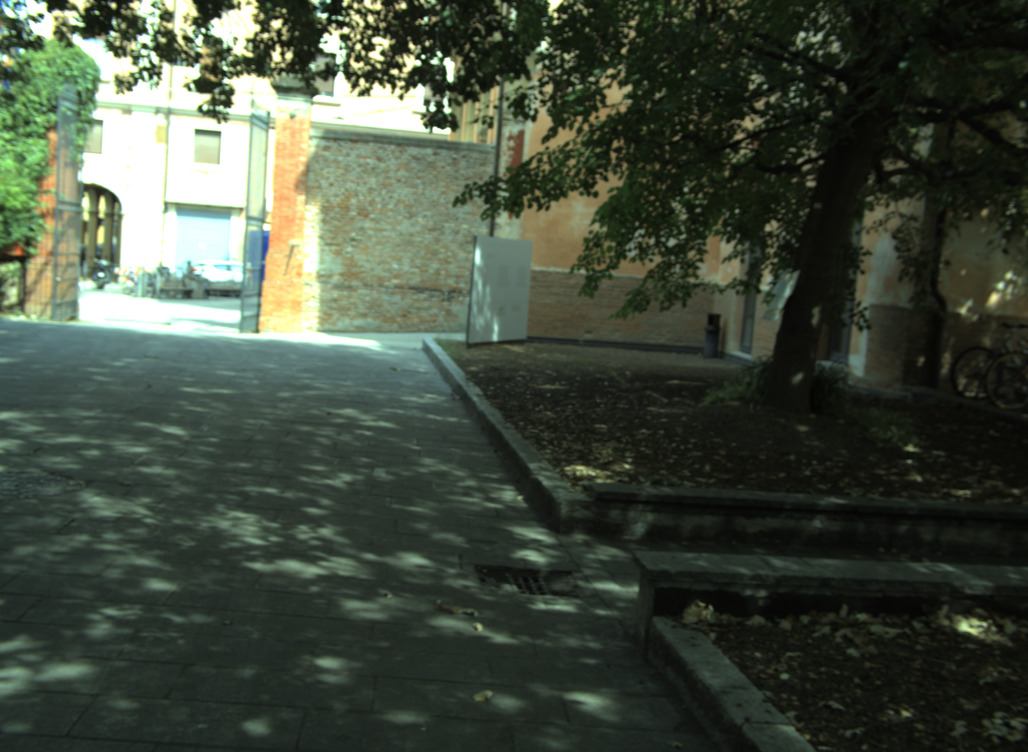} 
        {\colorbox{white}{$\displaystyle\textcolor{black}{\text{(e)}}$}}
        \end{overpic} &
        \begin{overpic}[width=0.31\linewidth]{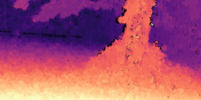} 
        {\colorbox{white}{$\displaystyle\textcolor{black}{\text{(f)}}$}}
        \end{overpic} \\
    \end{tabular}
    \vspace{-0.2cm}\caption{\textbf{Proxy Labels Distillation.} A classical algorithm \cite{hirschmuller2007evaluation} struggles at dealing with RGB-MS images (a-c). Using a second RGB camera yields much more accurate proxy labels (d-f). }
    \label{fig:proxies}
\end{figure}

\subsection{Proxy Label Distillation} We rely on proxy label distillation \cite{Tonioni_2017_ICCV,Tosi_CVPR_2019,aleotti2020reversing,Poggi2021continual} to obtain a strong and reliable source of supervision for our architecture from the unlabeled images. Thus, following \cite{Tosi_CVPR_2019}, we adopt the popular Semi-Global Matching (SGM) algorithm to obtain proxy-labels  from single-channel frames encoding the average of the image channels for the multi-spectral images and the converted grayscale for the RGB images. However, since SGM aims at computing matches between the two images, in our RGB-MS setup it is intrinsically tampered by the different modalities of the two, as shown in Fig. \ref{fig:proxies} (top).
This leads to less accurate proxy-labels and, thus, to a less effective weakly-supervised training. 

Thus, to source a reliable supervision also on the challenging RGB-MS setup, we propose to enhance our proxy distillation pipeline according to a novel strategy. Specifically, we have already discussed that a second high-resolution RGB image allows us to obtain accurate semi-dense ground-truth maps by means of the space-time stereo framework.
We argue that, in a similar manner, we can deploy the second RGB camera during unlabeled data acquisition as well to obtain better proxy supervision. This allows for complementing  any RGB-MS pair  with a corresponding RGB-RGB pair over which classical stereo algorithms can produce much more accurate proxy labels to be deployed at training time.
Hence, we exploit the very same procedure described before to warp the proxy-labels from the rectified RGB-RGB left image to the rectified RGB-MS left image. This second strategy, illustrated in Fig. \ref{fig:proxies} (bottom), allows us to supervise our network and teach it how to match RGB-MS modalities supervised by established knowledge concerning the matching between RGB-RGB, at the cost of requiring an additional RGB camera during the \textit{offline} self-training. This kind of strategy as already been exploited to pursue training without ground-truth depth labels, e.g.  in the case of monocular depth estimation by means of image reconstruction losses \cite{Godard_CVPR_2017,Godard_ICCV_2019} or to distill proxy labels \cite{Tosi_CVPR_2019}. 
Despite representing an additional requirement for the acquisition setup -- yet cheap and effortless compared to any annotation pipeline, such as the one used to collect our ground-truth data -- we will show in our experiments how this strategy, when such an additional sensor is available, leads to a more effective training of our RGB-MS and, thus, to more accurate disparity estimations.

\section{Experimental Results}

\subsection{Additional  Datasets}
\label{sec:adddata}
\textbf{UnrealStereo4K:} The UnrealStereo4K dataset \cite{Tosi2021CVPR} is a synthetic RGB-RGB stereo dataset containing around $8K$ high-resolution ($3840 \times 2160$) stereo pairs  with dense ground-truth disparities. Although it does not include multi-spectral images,  we employ UnrealStereo4K as an additional source of training data to improve the overall final results yielded by our cross-modal networks. In particular, as it is a standard RGB-RGB dataset, we simulate a RGB-MS setup by simply converting the right image of the stereo pair from RGB to grayscale and, then, stacking it to form a volume of 10 channels. Moreover, we resize input images fed to $\Psi$ using a downsample factor of $6$ to mimic the unbalanced factor featured by our real RGB-MS setup.

\subsection{Implementation Details}  
\label{sec:impdetails}
Our proposed architecture is implemented with PyTorch. We adopt Adam \cite{kingma2015adam} as optimizer, with $\beta_1= 0.9$  and $\beta_2= 0.999$. In our implementation, we adopt different stereo backbones for our $\Psi_\theta$ submodule. In particular, we conduct experiments on two different 3D architectures such as PSM \cite{chang2018psmnet} and GWC \cite{guo2019group}, showing that our proposal can be effectively combined with diverse stereo networks. 
For both the selected backbones, we use the official code provided by the authors and modify the input channels of the feature extractor associated with  the right image to match the number of bands of our multi-spectral camera (i.e., 10). 
For MLP$_C$ and MLP$_R$, instead, we follow the implementation of \cite{aleotti2021neural}.
We train each network for 70 epochs on a single NVIDIA 3090 GPU with a learning rate of $10^{-4}$.
As RGB-MS training set we use the whole set of 11K unlabeled images, computing SGM proxy labels on each stereo pair. 
We discard proxy disparity maps with less than 70\% of valid pixels to avoid too sparse supervision, obtaining approximately 9K valid training images.
When training also on UnrealStereo4K, we first pre-train for 30 epochs solely on synthetic data and then we train for other 70 epochs with both real and synthetic data. We empirically noticed that this training procedure produces sharper disparity maps as final results compared to training on RGB-MS pairs only.
During training we use a batch size of 2, random crops of size $1152 \times 1152$ from $L$ and select the corresponding crop of size $192 \times 192$ on the multi-spectral image $R$. Moreover, we randomly sample $P=30000$ points from each crop at continuous spatial locations in the image domain to pursue training based on  our continuous formulation. Additional training details are reported in the supplementary material.

\begin{table}[t]
    \centering
    \scalebox{0.6}{
    \begin{tabular}{l|c|cc|cc}
    \toprule
    Method & Input & $2^{\circ}$ RGB & Synth &  D-AEPE & ADE (m) \\
    \midrule
        SGM & Single-Channel & & & 14.66 & 0.215 \\
        Zhi et al. \cite{zhi2018deep} (photo) & Single-Channel &  & & 89.86 & 1.010 \\
        Zhi et al. \cite{zhi2018deep} & Single-Channel & \checkmark & & 19.45 & 0.116 \\
        \midrule
        PSM & All-Channels & & & 14.51 & 0.103 \\
        PSM & All-Channels & \checkmark & & 10.04 & 0.084 \\
        PSM + MLPs & All-Channels & \checkmark & & 19.12 & 0.182 \\
        PSM + MLPs + $\Phi$ & All-Channels & \checkmark &  & \textbf{8.36} & \textbf{0.065} \\
        \midrule
        PSM + MLPs + $\Phi$ & Single-Channel & \checkmark & & 24.13 & 0.737 \\
        \midrule
        PSM + MLPs + $\Phi$ & All-Channels & \checkmark & \checkmark & \textbf{7.31} & \textbf{0.057} \\	
        GWC + MLPs + $\Phi$ & All-Channels & \checkmark & \checkmark & 8.67 & 0.100 \\
    \bottomrule
    \end{tabular}}
    \vspace{-0.2cm}\caption{\textbf{Disparity/Depth Errors}. Quantitative results on the proposed RGB-MS dataset. We report error metrics for the disparity/depth predictions. Best results in bold.}
    \label{tab:results_disp}
\end{table}

\subsection{Evaluation Metrics}
To evaluate our RGB-MS model,  we use several metrics for optical flow, disparity and depth  estimation, computing all the metrics at high resolution. In case of the optical flow metrics, we consider the registration error from the RGB to the MS camera, which, indeed, is the performance metric most relevant in our scenario which, in essence, deals with enriching the RGB image with MS information taken from the available lower resolution sensor. Hence, the metrics used to evaluate the performance of our models are: 1) \textit{Flow Average End Point Error (F-AEPE)}, defined as the norm of the difference between the ground-truth and predicted flow vectors expressed in pixels, 2) \textit{Disparity Average End-Point-Error (D-AEPE)}, measured in pixels between the predicted disparity map and the ground-truth disparity map, averaged across all pixels, 3) \textit{Absolute depth error} (ADE) measured in meters (m), between the predicted depth map and the ground-truth depth map and 4) \textit{Percentage of bad pixels (bad$_\tau$)}, where $\tau$ is a tolerance on the optical flow error to accept an estimated flow as correct.

\begin{table}[t]
    \centering
    \scalebox{0.6}{
    \begin{tabular}{l|c|cc|cccccc}
    \toprule
    Method & Input & $2^{\circ}$ RGB & Synth & F-AEPE & bad$_1$ & bad$_2$ & bad$_3$ \\
    \midrule
        SGM & Single-Channel & & & 2.32 & 52.40 & 25.73	& 14.28 \\
        Zhi et al. \cite{zhi2018deep} (photo) & Single-Channel &  & & 14.22 & 85.05 & 73.87 & 66.42 \\
        Zhi et al. \cite{zhi2018deep} & Single-Channel & \checkmark & & 3.08 & 70.51 & 47.81 & 32.65 \\
        \midrule
        PSM & All-Channels & & & 2.30 & 44.49 & 21.42 & 13.07\\
        PSM & All-Channels & \checkmark & & 1.59 & 50.90 & 21.41 & 11.20\\
        PSM + MLPs & All-Channels & \checkmark & & 3.03 & 67.42 & 36.86 & 21.20\\
        PSM + MLPs + $\Phi$ & All-Channels & \checkmark &  & \textbf{1.32} & \textbf{34.89} & \textbf{13.32} & \textbf{7.05} \\
        \midrule
        PSM + MLPs + $\Phi$ & Single-Channel & \checkmark & & 3.82 & 73.20 & 43.08 & 22.20 \\
        \midrule
        PSM + MLPs + $\Phi$ & All-Channels & \checkmark & \checkmark &  \textbf{1.16} & \textbf{32.15} & \textbf{11.54} & \textbf{6.39} \\
        GWC + MLPs + $\Phi$ & All-Channels & \checkmark & \checkmark & 1.37 & 37.91 & 15.12 & 7.64 \\
    \bottomrule
    \end{tabular}}
    \vspace{-0.2cm}\caption{\textbf{Flow Errors}. Quantitative results on the proposed RGB-MS dataset. We report error metrics for the flow maps employed during registration. Best results in bold.}
    \label{tab:results_flow}
\end{table}

\subsection{Results}
We report experimental results and ablation studies dealing with  disparity/depth and flow error metrics in Tab. \ref{tab:results_disp} and Tab. \ref{tab:results_flow}, respectively.
As a reference, in the first row of both tables, we show the results yielded by SGM on the RGB-MS test set when both stereo frames are converted to grayscale, averaging all channels in case of the MS image. Moreover, we compare against \cite{zhi2018deep} by using their network on our RGB-MS stereo setup. 

\begin{figure*}[t]
    \centering
    \setlength{\tabcolsep}{1pt}
    \scalebox{1}{
    \begin{tabular}{ccccc}
         \textit{PSM} & \textit{+ 2$^\circ$ RGB} & \textit{+ MLPs} & \textit{+ $\Phi$} & \textit{+ Synth} \\
         \begin{overpic}[width=0.18\linewidth]{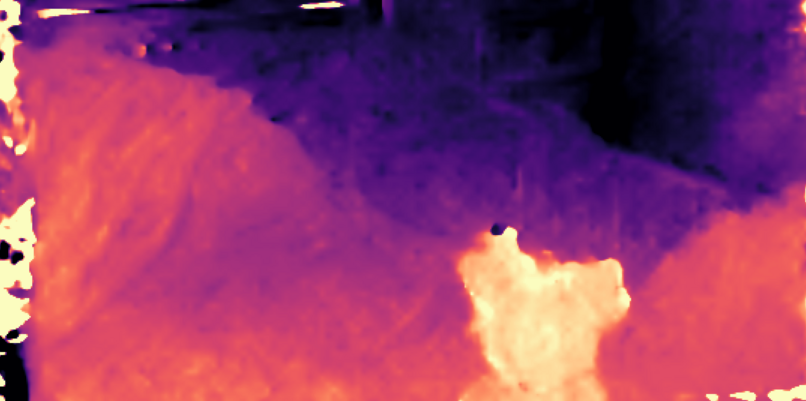}
         {\colorbox{white}{$\displaystyle\textcolor{black}{\text{(a)}}$}}
         \end{overpic} &
         \begin{overpic}[width=0.18\linewidth]{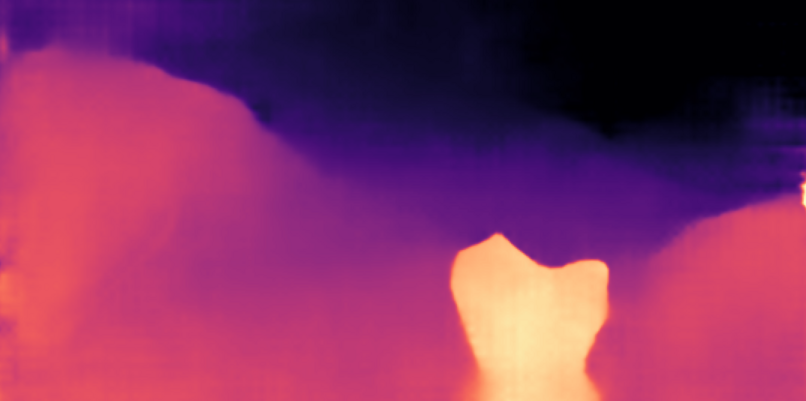} 
         {\colorbox{white}{$\displaystyle\textcolor{black}{\text{(b)}}$}}
         \end{overpic} &
         \begin{overpic}[width=0.18\linewidth]{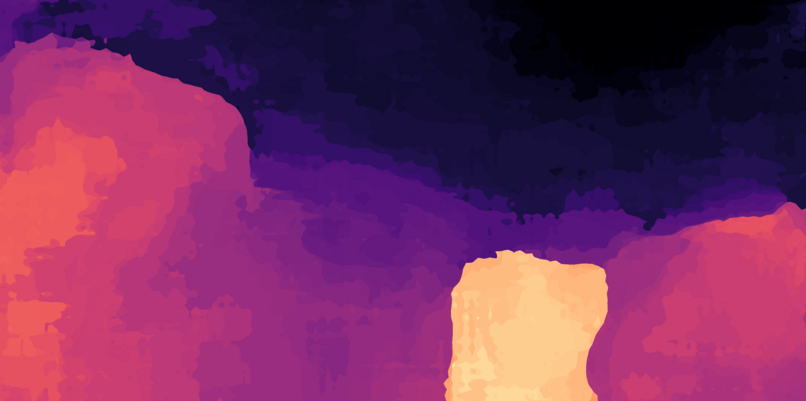} 
         {\colorbox{white}{$\displaystyle\textcolor{black}{\text{(c)}}$}}
         \end{overpic} & 
         \begin{overpic}[width=0.18\linewidth]{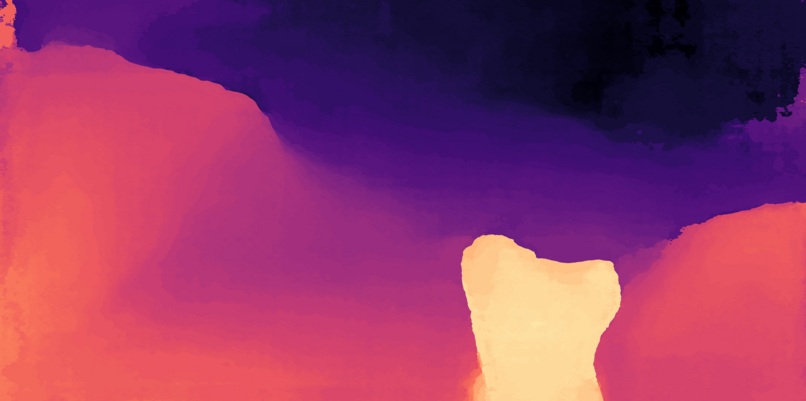} 
         {\colorbox{white}{$\displaystyle\textcolor{black}{\text{(d)}}$}}
         \end{overpic} & 
         \begin{overpic}[width=0.18\linewidth]{images/dataset/4_1/psm_hr_synth.png} 
         {\colorbox{white}{$\displaystyle\textcolor{black}{\text{(e)}}$}}
         \end{overpic} \\
    \end{tabular}}
    \vspace{-0.2cm}\caption{\textbf{Ablation Qualitative Results.} Disparity maps for the stereo pair in Fig. \hyperref[fig:teaser]{1} computed by: PSM (a), PSM with the proposed 2$^\circ$ RGB supervision (b), PSM with the continuous and sharp formulation leveraging the two MLPs (c). Finally, the full architecture with the high-resolution feature extractor $\Phi$ without (d) and with synthetic supervision (e).}
    \label{fig:qualitatives}
\end{figure*}

\textbf{Ablation on Supervision and Architecture.}
In rows from 4 to 7 of both tables, we ablate the contributions of the proposed supervision strategy and architectural components.
First of all, comparing the basic configuration of PSM trained on SGM proxy labels obtained with or without the adoption of the second additional RGB camera (rows 4 vs 5) we note an improvement for almost all metrics, with a sensible decrease of 4 pixels and almost 1 pixel in the D-AEPE and F-AEPE error metrics, respectively. 
These results support our idea that a better supervision obtained by two cameras with the same modality can improve the matching performance  across different modalities significantly.
In the sixth row, we add $MLP_C$ and $MLP_R$ to the PSM backbone, so as to introduce our continuous formulation potentially yielding much more accurate and sharp high-resolution disparity maps.  However, this causes a substantial deterioration of all performance metrics. We ascribe this issue to  the task  of recovering  an accurate  high-resolution disparity map through our continuous  formulation being too difficult in absence of any source of  high-resolution information. 
Indeed, in the following rows, we add  the $\Phi$ high-resolution feature extractor (Fig. \ref{fig:network}) to the architecture comprising the PSM backbone and the MLPs,   achieving the best results in all metrics with a huge boost compared to all previous configurations. 

We can appreciate the effect of each component in the first 4 columns of Fig. \ref{fig:qualitatives}. For instance, we can perceive how the continuous formulation based on the MLPs (third column)  can yield disparity maps exhibiting sharp edges, but only with the introduction of the encoder $\Phi$ (fourth column) we are able to recover correctly the high-resolution details. 

\textbf{Comparison with Existing Baselines.} In the second and third rows, we report the results obtained by \cite{zhi2018deep} by training it on our RGB-MS dataset, using the original photometric loss between the RGB image and the MS image averaged across channels, as well as on proxy labels extracted using the second RGB image. We can observe how, in both tables, the photometric loss (row 2) is in-effective  as the original formulation relies on photometrically calibrated RGB-NIR images and a mapping from RGB to pseudo-NIR, while the equivalent RGB to pseudo-MS mapping does not exist according to our MS camera manufacturer. On the contrary, by training it on proxy labels, it notably ameliorates the results -- although our network still represents the best solution for this task, since the architecture by \cite{zhi2018deep} is not designed to handle our unbalanced setup.

\textbf{Comparison with Alternative Input Modality.}
In row {8}, we perform an additional test comparing our proposal to a baseline strategy to handle different input modalities with a different number of channels,  -- i.e., collapsing both the left and right image  into a single-channel representation. We achieve this by converting RGB images into grayscale and averaging the 10 channels of MS images.
Thus, we can use the original formulation of the PSM architecture, which relies on a classical shared feature extractor. From Tab.\ref{tab:results_disp} and \ref{tab:results_flow} we notice how the network struggles with this concise input representation, suggesting that we better rely on the information from all bands to learn a reliable matching. 

\textbf{Auxiliary Synthetic Data.}
In both tables, we also report the results obtained by training the best architecture (rows 7) with additional  synthetic data (Sec. \ref{sec:adddata} and Sec. \ref{sec:impdetails} ). We note that we can further push its performances (rows 9 vs rows 7) achieving the best results in all metrics and with an average flow estimation error  (F-AEPE) as small as almost 1 pixel  (1.16), a remarkable result for such a challenging registration task dealing with images featuring diverse modalities and highly unbalanced resolutions. 
In the last column of Fig. \ref{fig:qualitatives} we can perceive qualitatively how leveraging on auxiliary synthetic training data can help to further improve  the predicted disparity maps, in particular in terms of sharpness of the finer details  thanks to the perfect ground-truth supervision in challenging regions such as depth discontinuities.

Eventually, in the last rows we report also the results obtained  using  a different backbone, such as GWC \cite{guo2019group}, proving to be effective as well.

\section{Conclusion, Limitations and Future work}

In this paper, we have explored RGB-MS image registration for the first time. Purposely, we have collected a dataset of 34 synchronized RGB-MS  pairs annotated with accurate semi-dense ground-truth disparity maps.
We have also proposed a deep network architecture to perform  cross-modal matching between the two image modalities. Moreover, to avoid the need of hard-to-source  ground-truth labels, we have proposed a novel supervision strategy aimed at  distilling knowledge from an easier  RGB-RGB matching problem. This strategy has been realized  by training our network on 11K unlabeled RGB-MS/RGB-RGB image pairs acquired  as part of our dataset.  
Testing the trained model on the annotated RGB-MS pairs shows promising results, with an average registration error close to 1 pixel despite the dramatic diversity in image modality and resolution. 

Although  our dataset is the first to address RGB-MS matching problem, it has several limitations. In particular, although comparable  in terms of  number of samples to some popular stereo datasets \cite{scharstein2014high}, more images could be collected to make our benchmark more complete and challenging. Moreover, at the moment, ground-truths are limited to indoor scenes only. Finally, images are acquired with only one specific 10-bands MS camera. 
As future work, we plan to  increase the size and variety of the dataset by acquiring more environments and employing diverse MS devices. 

Many other ideas should  definitely be explored to better  tackle the
novel and challenging registration task addressed in this paper. In this respect,  we believe that
our proposed architecture and training methodology set forth a  strong baseline  to  steer further investigation  towards more effective and more general solutions.

\textbf{Acknowledgements.} We gratefully acknowledge the funding support of Huawei Technologies Oy (Finland). 

{\small
\bibliographystyle{ieee_fullname}
\bibliography{egbib}
}

\newpage\phantom{Supplementary}
\multido{\i=1+1}{6}{


\section*{Supplementary material}

# Supplementary Material for
# RGB-Multispectral Matching: Dataset, Learning Methodology, Evaluation

Fabio Tosi*   Pierluigi Zama Ramirez*   Matteo Poggi*
Samuele Salti   Stefano Mattoccia   Luigi Di Stefano
CVLAB, Department of Computer Science and Engineering (DISI)
University of Bologna, Italy
{fabio.tosi5, pierluigi.zama, m.poggi}@unibo.it

In this document, we provide additional details concerning CVPR submission "RGB-Multispectral Registration: Dataset, Learning Methodology, Evaluation".

## 1. Calibration of our trinocular rig

In this section, we detail the calibration procedure needed to gather images and either i) annotate them with accurate ground-truth depth labels or ii) distill accurate proxy labels.

We define the three cameras of our setup as $RGB$, $MS$, and $RGB^2$, with $RGB$ and $RGB^2$ being the two RGB cameras with identical resolution ($W_{RGB} \times H_{RGB}$) and $MS$ the multi-spectral lower resolution device ($W_{MS} \times H_{MS}$). We define raw images acquired by the $RGB$, $MS$, and $RGB^2$ cameras as $rgb$, $ms$, and $rgb^2$ respectively.

First of all, we calibrate each camera independently, using the standard calibration procedure from OpenCV based on the detection of corners on a chessboard pattern. We found out empirically that we could detect corners with standard algorithms also in our MS images, simply by processing a grayscale image defined as the average across channels. This allows for straightforward calibration of the MS camera.

Then, following again the standard procedure from OpenCV, we calibrate the $RGB - RGB^2$ stereo system and rectify images. In this way we obtain rectified images $rgb_2, rgb_2^2$ for the $RGB - RGB^2$ stereo system.

Finally, we calibrate the *unbalanced* stereo system $RGB - MS$ following the procedure discussed below.

**Unbalanced Stereo Rectification.** To rectify images acquired by the $RGB - MS$ unbalanced stereo system, we follow the *unbalanced rectification* scheme sketched in [1], yielding images that are rectified when brought to the same resolution by means of up-sampling or down-sampling operations solely.

We denote the camera with the smaller $HFOV$ as $j$ while the other one as $i$.

$$\begin{cases} i = RGB, j = MS & \text{if } HFOV_{MS} < HFOV_{RGB} \\ i = MS, j = RGB & \text{if } HFOV_{RGB} < HFOV_{MS} \end{cases} \quad (1)$$

By modifying the intrinsic parameters of $i$, we simulate a crop and scale change so as to match the $HFOV$, Aspect Ratio ($AR$) and size of $j$. Then, we compute the rectification transformation baased on these new parameters.

Hence, we compute the new width and height of $i$, $\hat{W}_i$ and $\hat{H}_i$, which we use to crop the image with the larger $HFOV$ so as to match the smaller $HFOV$ one while preserving the aspect ratio

$$\hat{W}_i = 2 \tan \frac{HFOV_j}{2} f_i \quad (2)$$

$$\hat{H}_i = \frac{H_j}{W_j} \hat{W}_i \quad (3)$$

Then, we change the intrinsic parameters of $i$ to simulate the crop and resize, and can thus match the resolution of $j$ as follows:

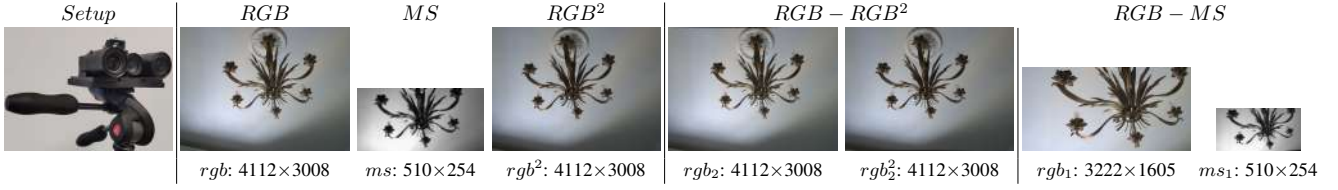

Figure 1. **Rectification example.** From left to right: our trinocular rig, $rgb$, $rgb^2$, and $ms$ raw images acquired by our rig. $rgb_2$, $rgb_2^2$ rectified balanced stereo pair from of the $RGB - RGB^2$ stereo system. $rgb_1, ms_1$ unbalanced rectified stereo pair from of the $RGB - MS$ stereo system.

$$\hat{A}_i = \begin{bmatrix} f_x^i \cdot \frac{W_j}{\hat{W}_i} & 0 & (u_0^i - \frac{W_i - \hat{W}_i}{2}) \cdot \frac{W_j}{\hat{W}_i} \\ 0 & f_y^i \cdot \frac{H_j}{\hat{H}_i} & (v_0^i - \frac{H_i - \hat{H}_i}{2}) \cdot \frac{H_j}{\hat{H}_i} \\ 0 & 0 & 1 \end{bmatrix}$$

Then, we estimate the rectification transformation as we would have two cameras of height $H_j$ and width $W_j$, finding the new intrinsics and rotations to map the initial image plane into the rectified one. As we have estimated the intrinsic matrices at the resolution of $j$, we rescale the intrinsic matrix of $i$ with a vertical and horizontal scale factors equal to $\frac{\hat{H}_i}{H_j}$ and $\frac{\hat{W}_i}{W_j}$, respectively, in order to adjust the focal length and piercing point of the camera.

Finally, we can rectify the unbalanced pair obtaining two rectified images $rgb_1, ms_1$ for the $RGB - MS$ stereo system.

Fig. 1 shows an example of images before and after the rectification procedures described so far.

## 2. Disparity Warping

When using $RGB^2$, both during the proxy labels creation or the ground-truth acquisition, we need to warp the left ground-truth $Disp_2$ aligned with $rgb_2$ of the $RGB - RGB^2$ stereo system in order to obtain the ground-truth $Disp_1$ aligned with the $rgb_1$ of the $RGB - MS$ stereo system.

We know that the rectification transformation is only a change of intrinsic parameters and a rotation, thus an homography. Therefore, we can compute the mapping between pixels of $rgb_2$ of the $RGB - RGB^2$ stereo system, with coordinate $(u, v)$, and pixels of the left image $rgb_1$ of the $RGB - MS$ stereo system, with coordinate $(u', v')$ as:

$$\begin{pmatrix} u' \\ v' \\ 1' \end{pmatrix} = A_1 R_1 R_2^{-1} A_2^{-1} \begin{pmatrix} u \\ v \\ 1 \end{pmatrix} \quad (4)$$

where $A_1, R_1$ and $A_2, R_2$ are the intrinsic and rectification rotation matrix of the camera $RGB$ in the $RGB - MS$ and $RGB - RGB^2$ stereo systems, respectively.

Once this mapping is known, we can perform a backward warping to obtain $Disp_1$ from $Disp_2$. However, we need to modify the disparity values according to the 3D rotation and baseline change before warping. Thus, given the disparity map $Disp_2$, we first transform it into the corresponding depth map $D_2$

$$D_2 = \frac{f_2 b_2}{Disp_2} \quad (5)$$

where $f_2$ is the focal length of the rectified $rgb_2$ and $b_2$ is the baseline of the stereo system $RGB - RGB^2$. Then, we back-project each pixel of $rgb_2$ into the 3D space using $D_2$ and we rotate it, obtaining the pixel in the $rgb_1$ reference frame:

$$\begin{pmatrix} x' \\ y' \\ z' \end{pmatrix} = R_1 R_2^{-1} D_2 A_2^{-1} \begin{pmatrix} u \\ v \\ 1 \end{pmatrix} \quad (6)$$

In this way we can create a depth map $D_{2\to1}$ for which any pixel $(u, v)$ contains the depth value of the corresponding pixel aligned in the $rgb_1$ reference frame, $z'$. At this point we perform the backward warping of the depth:

$$D_1 = \phi(D_{2\to1}) \quad (7)$$

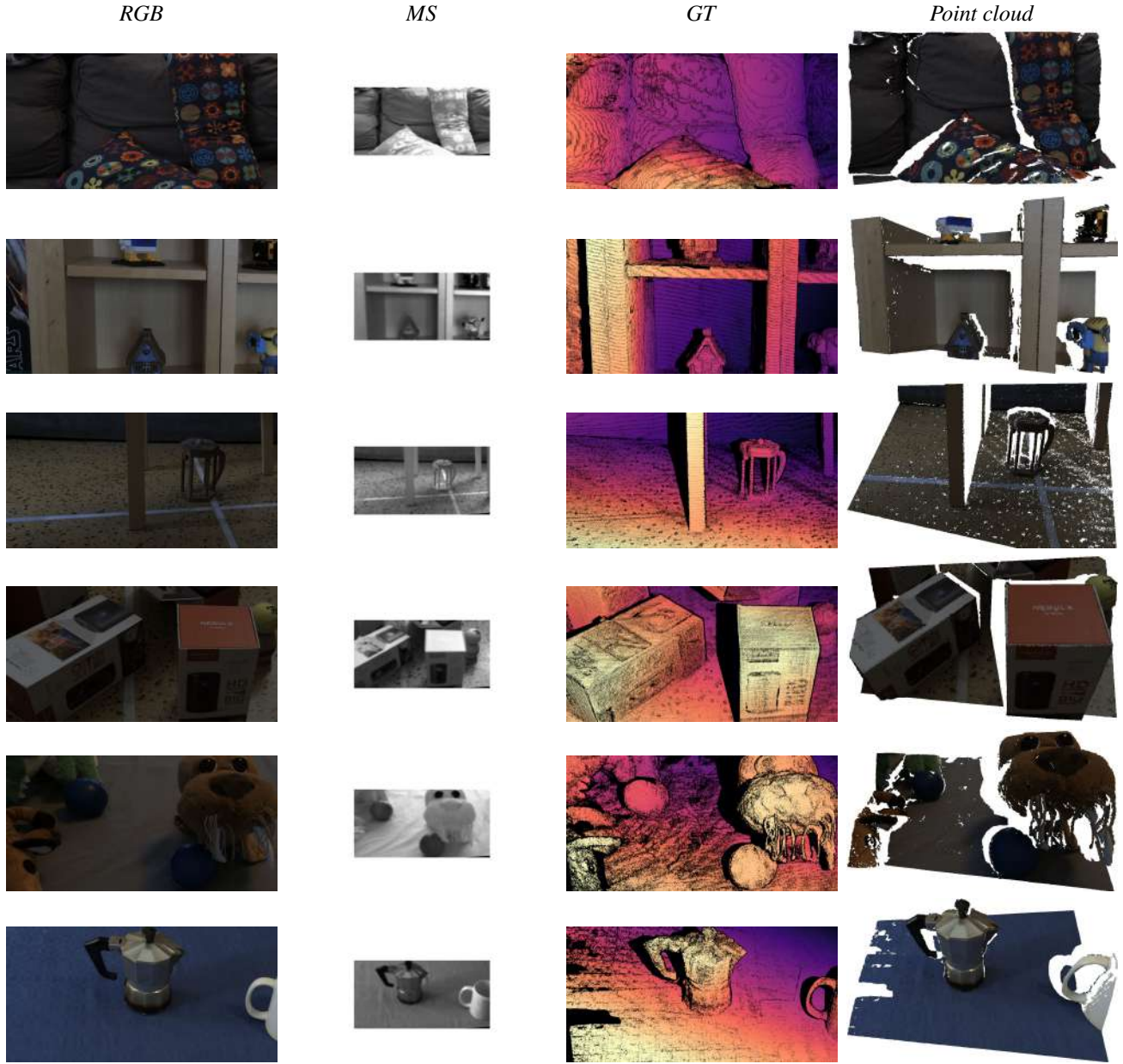

Figure 2. **Ground Truth Examples.** From left to right: RGB image, MS image (mean across channels), ground-truth disparity map obtained with our acquisition pipeline, point cloud visualization.

where $\phi$ is the backward warping operation that use the mapping defined at Eq. 4 and $D_1$ is the depth map aligned with $rgb_1$. Finally we transform it to the ground disparity map of $rgb_1$ as:

$$Disp_1 = \frac{f_1 b_1}{D_1} \qquad (8)$$

where $f_1$ and $b_1$ are the focal length of $rgb_1$ and the baseline of the $RGB - MS$ stereo system.

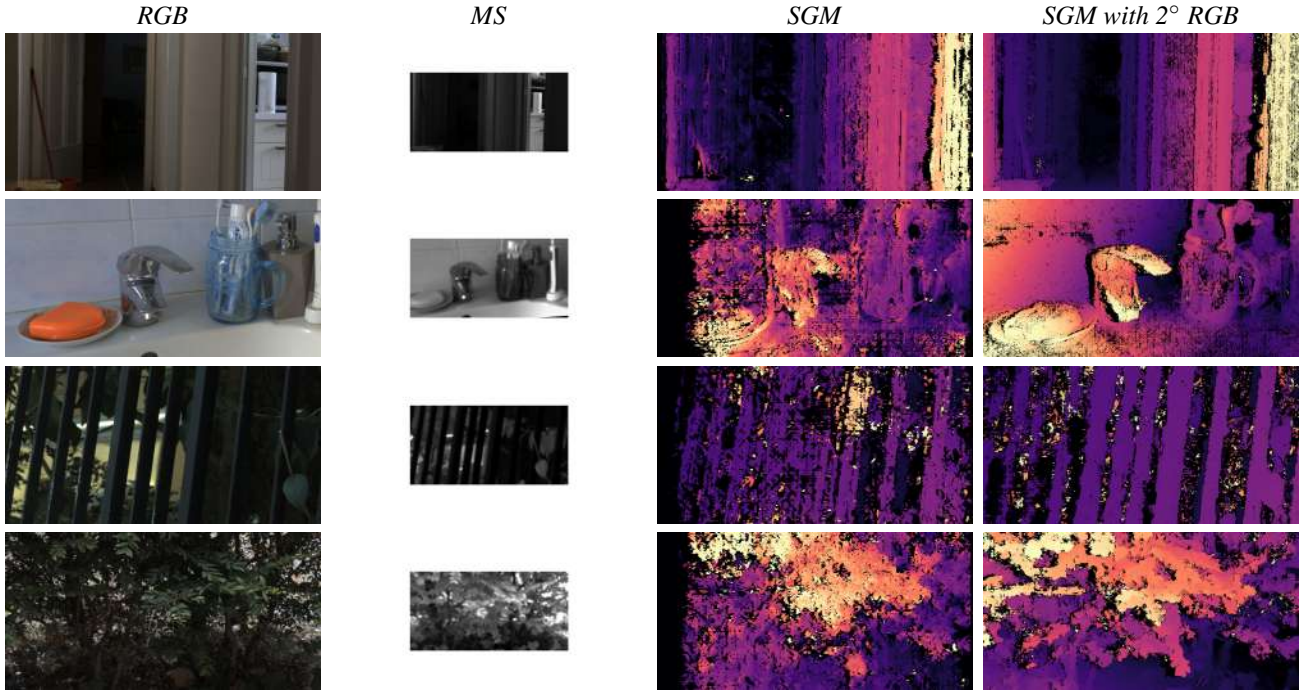

Figure 3. **Unlabeled Examples.** From left to right: RGB image, MS image (mean across channels), SGM computed between grayscale representation of RGB and MS images, SGM computed on two RGB cameras and then warped.

## 3. Additional Qualitative Results

In this section, we present additional qualitative results. In Fig. 2 we show examples of RGB-MS pairs with available disparity ground-truths. In Fig. 3 we present examples of the images acquired in uncontrolled environments used to train our deep architecture by our proxy supervision approach. In particular, in the last two columns we show SGM proxy labels computed on a RGB image converted into grayscale and a single channel MS image attained by computing the mean across all channels at each pixel, as well as the SGM proxy labels obtained by two RGB images and then warped. Finally, in Fig. 4 we report additional qualitative results obtained by our network using PSM as backbone and trained also with auxiliary synthetic data (i.e., the best configuration of Tab. 1 and Tab 2. of the main paper).

## 4. Qualitative Registration Results.

To provide hints on the registration quality, in Fig. 5 we report qualitatives obtained by using either the disparity predicted by our network or the ground-truth to warp the low resolution MS image into the high-resolution RGB image. We notice how the two warped images appear very similar, which vouches for the effectiveness of our cross-modal registration architecture.

## 5. Additional Implementation Details

We provide here additional details regarding our deep cross-spectral network architecture as well as on the adopted continuous output representation.

### 5.1. Stereo Backbones

As for the stereo backbones, we follow the original implementation of two popular architectures: PSM [2] and GWC [3]. For both of them, we follow the idea proposed in [4] to exploit a different combination of feature maps computed at different spatial resolutions in order to capture both local and global context.

**PSM** [2]: For the PSM stereo backbone, we use the original deep feature extractor, consisting of Spatial Pyramid Pooling (SPP) modules as $\Phi_\theta$ while the stereo matching cost probabilities are extracted from stacked hourglass 3D convolutions on the cross-spectral cost volume computed by $\Psi_\theta$. More specifically, we perform bilinear interpolation on features of size

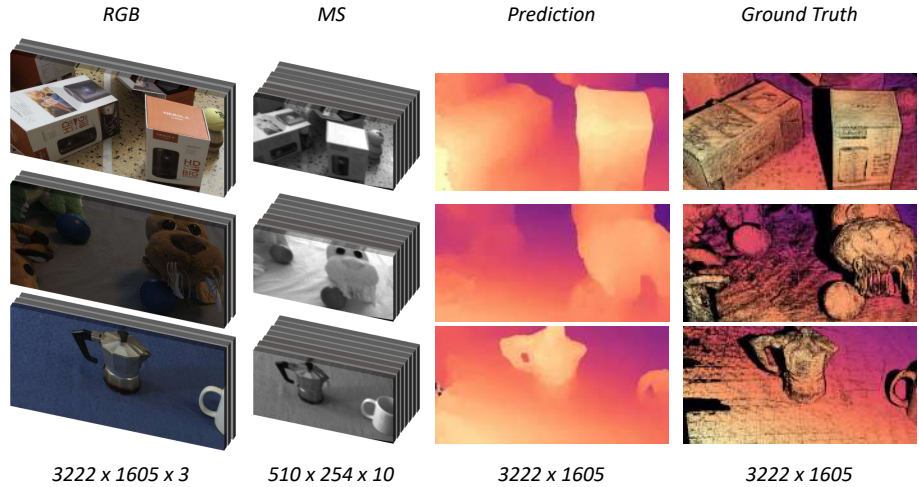

Figure 4. **Additional Qualitative Results of our Network**. From left to right: input RGB image, input multi-channels MS image, disparity predicted by our network, ground-truth.

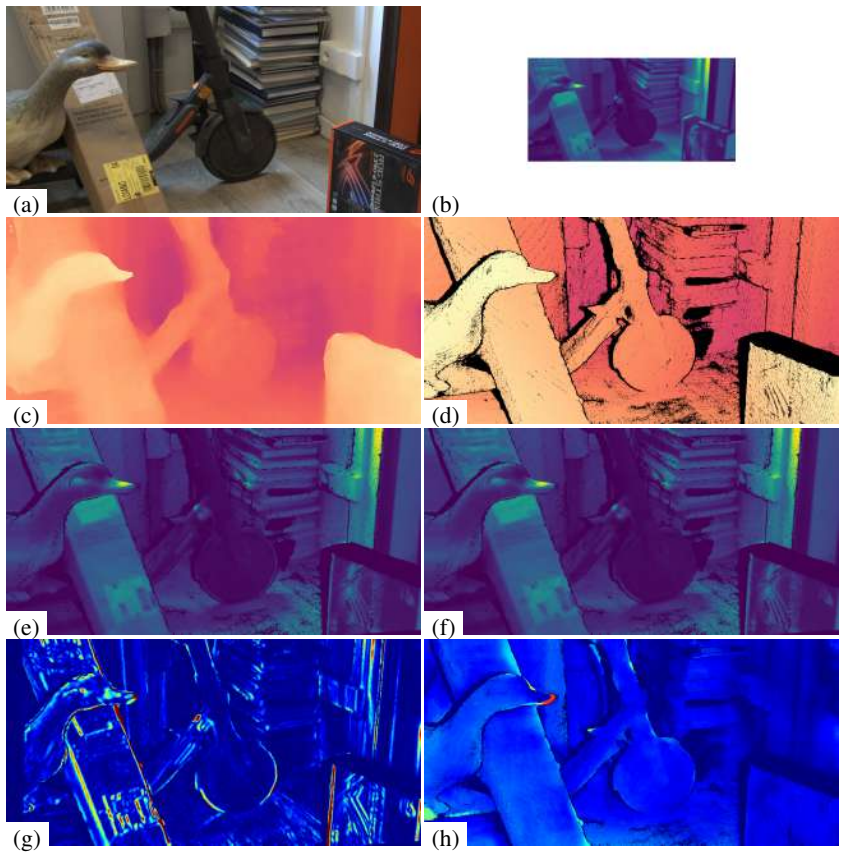

Figure 5. **Registration Results.** RGB image (a); MS image displayed as a channel average with color-map (b); predicted (c) and ground-truth (d) disparity maps; MS images warped based on the predicted (e) and ground-truth (f) disparities; absolute difference between registered images (g) and disparities (h).

$576 \times 960 \times 32$ computed by $\Phi_\theta$ and on matching cost probabilities of dimension $92 \times 160 \times 32$ computed by $\Psi_\theta$. It is worth noticing that, for the cost volume computation, we modify the original implementation of PSM such that the two feature

extractors for $L$ and $R$ do not share weights and accept 3 channels as input for the RGB branch whilst 10 channels for the multi-spectral branch.

**GWC** [3]: The GWC stereo backbone shares the same feature extractors as in PSM, while $\Psi_\theta$ incorporates group-wise correlation to build-up the cross-spectral cost volume. Specifically, we perform bilinear interpolation on convolutional features of size $576 \times 960 \times 332$ from $\Phi_\theta$ and matching cost probabilities of dimension $92 \times 160 \times 32$ from $\Psi_\theta$.

### 5.2. Continuous Output Representation

Our output representation follows the same implementation as proposed in [1], where two multi-layer perceptrons are in charge of estimating a categorical distribution over disparity values and a subpixel offset starting from interpolated features computed by the deep stereo backbone. More specifically, the number of neurons is $(D_\psi + F_\phi, 512, 256, 128, d_{max})$ for $MLP_C$ while $(D_\psi + F_\phi + 1, 128, 64, 1)$ for $MLP_R$, where $D_\psi = 32$ and $F_\phi = 32$ in case of the PSM stereo backbone while $D_\psi = 32$ and $F_\phi = 332$ for GWC. In our experiments, we always use $d_{max} = 768$. Notice both multi-layer perceptrons rely on Sine activation functions, except the last layer where Softmax and Tanh activations are used for $MLP_C$ and $MLP_R$, respectively.


}

\end{document}